\documentclass[10pt,twocolumn,letterpaper]{article}

\usepackage[pagenumbers]{cvpr} % To force page numbers, e.g. for an arXiv version

\usepackage{algorithm}
\usepackage{algorithmic}
\usepackage[normalem]{ulem}
\usepackage{amsmath} 
\usepackage{tikz}
\usetikzlibrary{decorations.pathreplacing, calc}
\usepackage{multirow}
\usepackage[table]{xcolor}
\usepackage{makecell}

\definecolor{cvprblue}{rgb}{0.21,0.49,0.74}

\renewcommand{\maketitlesupplementary}{
    \clearpage           
    \onecolumn           
    \setcounter{page}{1}

    \begin{center}
        \Large
        \textbf{\thetitle}\\
        \vspace{0.5em}Supplementary Material \\
        \vspace{1.0em}
    \end{center}
}
\usepackage[pagebackref,breaklinks,colorlinks,allcolors=cvprblue]{hyperref}

\title{FlowDC: Flow-Based Decoupling-Decay for Complex Image Editing}

\author{
    Yilei Jiang\textsuperscript{1,$*$}, 
    Zhen Wang\textsuperscript{2,$*$ $\dagger$},
    Yanghao Wang\textsuperscript{2}, 
    Jun Yu\textsuperscript{3}, 
    Yueting Zhuang\textsuperscript{1}, 
    Jun Xiao\textsuperscript{1}, 
    Long Chen\textsuperscript{2} \\
    \textsuperscript{1}Zhejiang University \quad 
    \textsuperscript{2}HKUST \quad
    \textsuperscript{3}Harbin Institute of Technology (Shenzhen) \\
    {\tt\small yileijiang@zju.edu.cn} \quad
    {\tt\small zhenwang@ust.hk} \quad
    {\tt\small ywangtg@connect.ust.hk} \\
    {\tt\small junx@cs.zju.edu.cn} \quad
    {\tt\small longchen@ust.hk} \quad
}

\begin{document}

\twocolumn[{%

\maketitle

\begin{center}
    \centering
    \captionsetup{type=figure}
    \vspace{-2em}
    \includegraphics[width=1.0\textwidth]{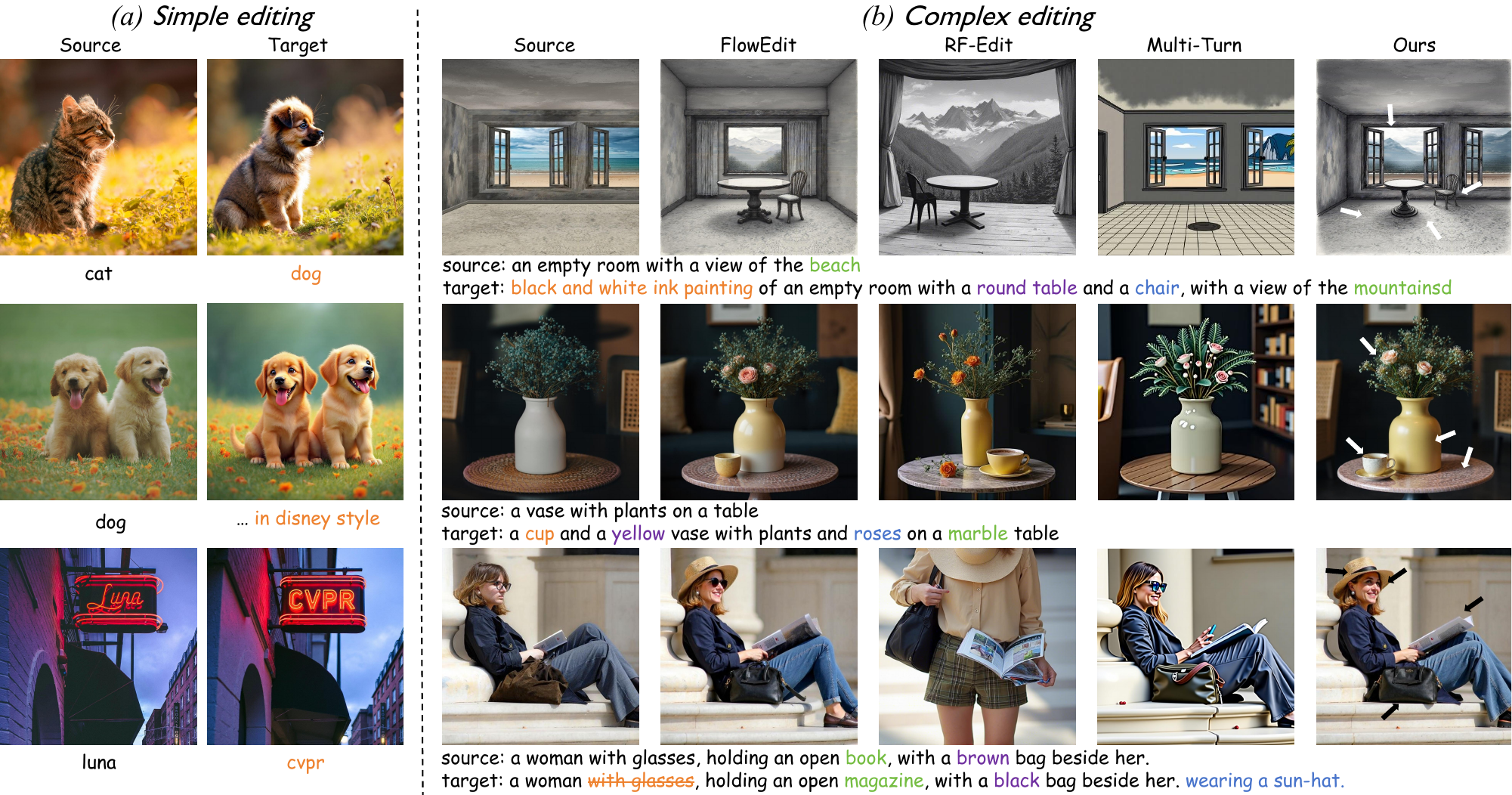}
    \vspace{-2em}
    \caption{
    (a) \textbf{Simple editing}: single editing target (modify one item). (b) \textbf{Complex editing}: Multiple editing targets (modify multiple items). Compared with existing methods, ours can achieve a good balance between semantic alignment and source consistency.}
    \label{fig:intro1}
\end{center}
}]

\begin{abstract}
With the surge of pre-trained text-to-image flow matching models, text-based image editing performance has gained remarkable improvement, especially for \underline{simple editing} that only contains a single editing target. To satisfy the exploding editing requirements, the \underline{complex editing} which contains multiple editing targets has posed as a more challenging task. However, current complex editing solutions: single-round and multi-round editing are limited by long text following and cumulative inconsistency, respectively. Thus, they struggle to strike a balance between semantic alignment and source consistency.
In this paper, we propose \textbf{FlowDC}, which decouples the complex editing into multiple sub-editing effects and superposes them in parallel during the editing process. Meanwhile, we observed that the velocity quantity that is orthogonal to the editing displacement harms the source structure preserving. Thus, we decompose the velocity and decay the orthogonal part for better source consistency.
To evaluate the effectiveness of complex editing settings, we construct a complex editing benchmark: Complex-PIE-Bench. On two benchmarks, FlowDC shows superior results compared with existing methods. We also detail the ablations of our module designs.
\end{abstract}  
\section{Introduction}
\label{sec:intro}
% The significance of complex editing
Flow matching~\cite{lipman2022flow,liu2022flow} models (FMs) have recently demonstrated remarkable generative ability to satisfy various generation requirements. Especially, pre-trained text-to-image FMs like~\cite{esser2024scaling,labs2025flux1kontextflowmatching} have been widely adapted for the task of text-based image editing, \ie, editing a source image under the guidance of a target prompt. While substantial progress has already been made with convincing editing performances, current researches~\cite{kulikov2025flowedit,kim2025flowalign,rf_inversion} mainly focus on the scenario of \textbf{\emph{simple editing}}. By ``simple", we refer to cases where the target prompt contains only a single editing target, \ie, modifying a specific item (\eg, an object or attribute) with a specific editing type (\eg, object removal or style change), as illustrated in Figure~\ref{fig:intro1}(a).

\begin{figure}[tb] 
    \centering
    \includegraphics[width=1\linewidth]{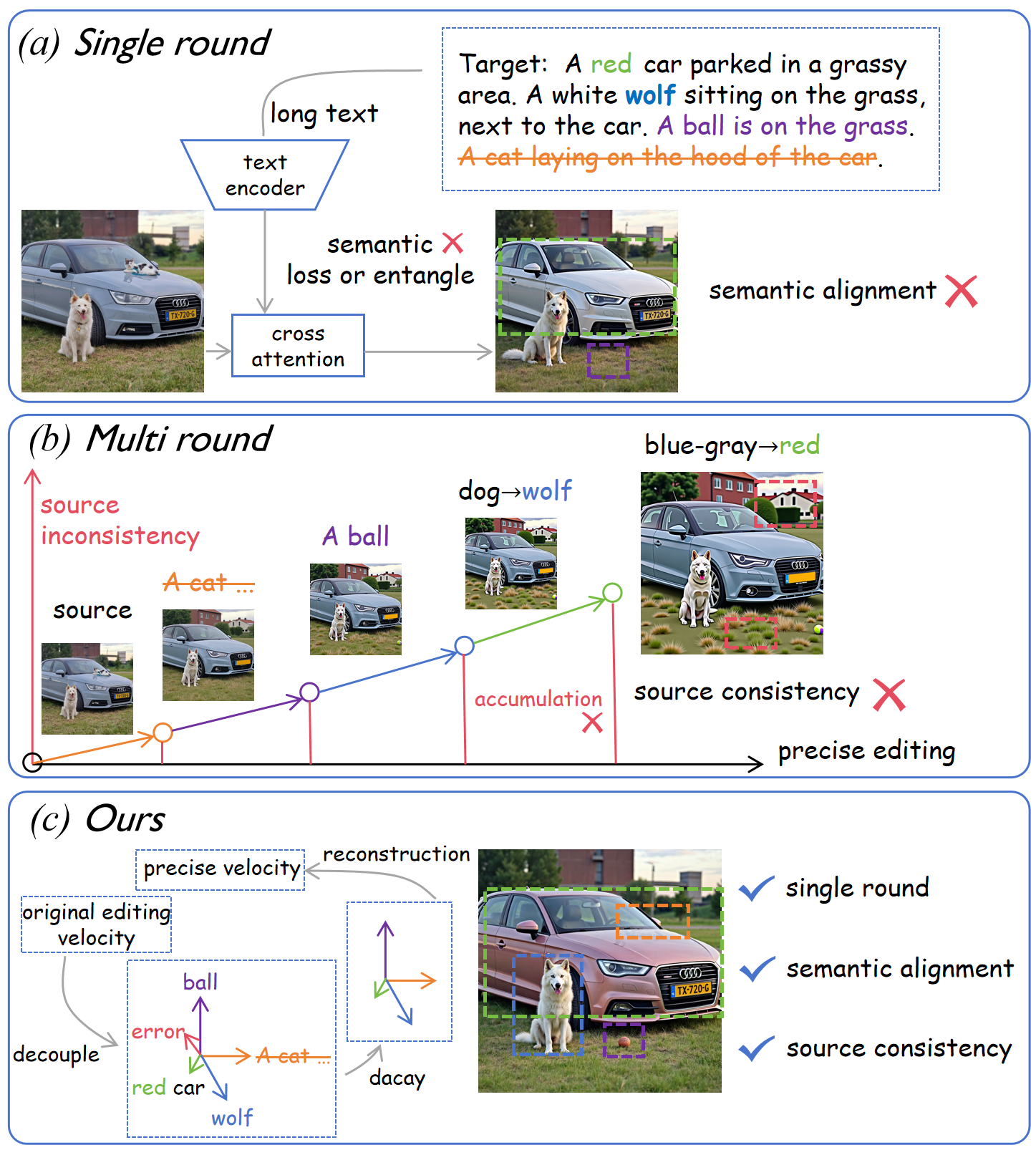}
    \caption{
    (a) Single-round editing methods struggle to handle long target prompts. 
    (b) Multi-round editing methods cause an accumulation of source inconsistency. 
    (c) Our method introduces velocity decoupling and decay techniques, achieving a balance between semantic alignment and source consistency.
    }
    \label{fig:intro_2}
    \vspace{-1em}
\end{figure}

However, the exploding demands from real-world applications increasingly require the ability to edit multiple contents within an image, a capability that has recently begun to attract growing attention~\cite{zhou2025multi,guo2024focus,swami2025promptartisan}. Thus, in this paper, we focus on the more challenging and practical problem of \textbf{\emph{complex editing}}, where the target prompt specifies multiple independent editing targets, each with its own item and type. As shown in Figure~\ref{fig:intro1}(b), compared with simple editing, complex editing introduces additional difficulties due to the coexistence of multiple editing targets. Specifically, it faces greater challenges over the two common goals of text-based image editing: 1) \textit{Semantic Alignment}: the synthesized image should faithfully follow all item changes that adhere to the multiple editing targets without omissions or confusion of each effect. 2) \textit{Source Consistency}: The synthetic image should maintain the editing-irrelevant parts the same as the source image. To achieve this, existing methods address complex editing through two primary paradigms:

1) \textbf{Single-round editing}: A straightforward solution is to directly apply existing text-based image editing methods~\cite{kulikov2025flowedit,kim2025flowalign,wang2025flowcycle} by treating the complex target prompt as a simple prompt, which ignores its complexity. As shown in Figure~\ref{fig:intro_2}(a), the complex prompt is directly fed into the FM and the final edited image is generated in a single round. However, today's pre-trained FMs are limited in their ability to handle long-text semantics. For example, when there are many editing targets in the long target prompt, the editing results usually lose or entangle some of the editing effects. 
Thus, follow-up methods~\cite{zhou2025multi, rf-solver} introduce attention manipulation to help locate editing areas. Even so, the editing performance still struggles due to problems such as mask overlap and limited attention generalization.

2) \textbf{Multi-round editing}: To avoid inputting the long text into the FMs, some studies~\cite{zhou2025multi,guo2024focus,swami2025promptartisan} decompose the long target prompt into a set of short texts (\eg, each prompt only contains one editing target). Then it iteratively applies the simple editing processes for multiple rounds to replace the single complex editing. Each round is only responsible for one editing target. In that case, the overall complex editing can be accomplished by superposing the multi-round simple editing effects. However, the multi-round editing is not only time-consuming (the overhead is linearly positively correlated with the round number), but also leads to unbearable cumulative source inconsistency (\cf, Figure~\ref{fig:intro_2}(b)). 

In this paper, we argue that all existing methods struggle in achieving a good balance between the semantic alignment and source consistency. To this end, we propose a simple yet effective \textbf{\underline{Flow}} \textbf{\underline{D}}e\textbf{\underline{c}}oupling and \textbf{\underline{D}}e\textbf{\underline{c}}ay method for complex image editing: \textbf{FlowDC}, which can faithfully react to each editing target in the complex text, as well as maintain high source consistency. 
Intuitively, we decouple the complex text into multiple progressive texts, each of which records the direction of specific editing targets. 
During the editing process, the complex velocity will be projected to a subspace spanned by decoupled text's directions and then reconstructed after decaying its orthogonal velocity component (\cf, Figure~\ref{fig:intro_2}(c)). 
In that case, the in-subspace velocity component captures the effect of each editing target, while the orthogonal component usually corresponds to unstable, editing-irrelevant structural changes, prompting its decay to yield a more precise velocity during reconstruction.

Specifically, our method contains two designs: 
1) \textbf{Progressive Semantic Orthogonalization.} 
We first decouple the complex target prompt into a series of intermediate target prompts, which are used to construct multiple editing trajectories. The resulting set of editing directions is then decoupled into an orthogonal basis, each isolating the semantic contribution of a specific editing target.
2)\textbf{Velocity Orthogonal Decay.} 
To reconstruct more precise velocity, the original editing velocity is projected onto a time-varying subspace, which is spanned by the orthogonal basis obtained in PSO (or alternatively, via displacement). 
During this reconstruction, the in-subspace component is primarily retained, while the orthogonal component strongly decays.
Combining these two designs ensures the reconstructed velocity focuses purely on the desired edits, simultaneously preserving source consistency and maintaining semantic alignment with the complex target prompt.

We summarize our contributions in threefold: 1) We analyzed the challenge of complex text-based image editing tasks and indicated the limitations of existing methods, \ie, long text following, expensive overhead, and accumulative source inconsistency. 2) We proposed \textbf{FlowDC} that decouples the complex editing into the superposition of multiple target editing effects and decays the orthogonal component. It can faithfully react to each editing target and preserve the source consistency. 3) We evaluated our method on an existing benchmark and, for more comprehensive validation, on a new complex editing dataset we constructed. The extensive results demonstrate the advancement of our method.

\section{Related Work}
\label{sec:related_work}

\noindent\textbf{Text-based Image Editing.} 
Current approaches can be broadly categorized into two main groups. 
The first is inversion-based editing~\cite{song2022denoisingdiffusionimplicitmodels, rf_inversion, rf-solver, mengsdedit, fireflow, mokady2023null, ju2023direct, miyake2025negative, zhou2025multi, deutch2024turboedit, tumanyan2023plug, tsaban2023ledits, wallace2023edict, wu2023latent, samuellightning, huberman2024edit, brack2024ledits++, wang2025flowcyclepursuingcycleconsistentflows}.
This popular paradigm typically involves a two-stage process: first inverting the source image into a noise latent using techniques like DDIM Inversion~\cite{song2022denoisingdiffusionimplicitmodels}, and then sampling from this latent, guided by a target prompt, to synthesize the edited image.
A second group of approaches, such as FlowEdit~\cite{kulikov2025flowedit} and FlowAlign~\cite{kim2025flowalign}, uses inversion-free methods for directly transforming a source image into a target image. By often requiring a lower transport cost (e.g., a shorter, more direct path in the probability flow), these methods have a better performance at preserving source consistency.
% A second group of approaches, such as FlowEdit~\cite{kulikov2025flowedit} and FlowAlign~\cite{kim2025flowalign}, uses inversion-free methods for directly transforming the source image into the target image. 
To better preserve the original image structure while enhancing editing strength, many methods \cite{hertz2022prompttopromptimageeditingcross, chung2024style, huang2024paralleledits, zhou2025multi, rf-solver, cao2023masactrl, couairon2022diffedit, dkm2025improving, Avrahami_2025_CVPR} manipulate attention maps during the synthesis process. Techniques like Prompt-to-Prompt~\cite{hertz2022prompttopromptimageeditingcross} achieve this by replacing or refining cross-attention maps to control the spatial layout and influence of text tokens. This technique has been particularly adapted by~\cite{huang2024paralleledits, zhou2025multi, dkm2025improving} for complex editing tasks, which are more challenging than simple editing. However, these attention-based methods struggle when edits are global or semantically overlapping, and may require careful prompt engineering.

\noindent\textbf{Multi-round Image Editing.} 
Current research achieve complex editing tasks by editing in multiple rounds. 
Multi-turn~\cite{zhou2025multi} employs a Linear Quadratic Regulator and adaptive attention manipulation during iterative editing rounds. 
Separately, Kim et al.~\cite{dkm2025improving} proposed a layer-wise memory framework to maintain consistency across multiple rounds.
EMILIE~\cite{joseph2024iterative} introduces a  latent iteration strategy based on IP2P~\cite{brooks2023instructpix2pix}, which takes intsructions instead of text prompts as inputs.
% Other methods~\cite{cui2023chatedit, li2024textbind, yang2023idea2img} utilize LLMs to support interactive image editing. 
Other methods leverage LLMs to manage more complex interactions, such as tracking dialogue history~\cite{cui2023chatedit}, synthesizing instruction data~\cite{li2024textbind}, or enabling iterative self-refinement~\cite{yang2023idea2img}.
Unlike these prior works, our proposed method is tailored for text-based, complex editing tasks in a single round without attention manipulation.

\section{Method}
\label{sec:method}

\subsection{Preliminary}

\noindent\textbf{Rectified Flow.}
Rectified Flow~\cite{liu2022flow} learns an Ordinary Differential Equation (ODE), 
$dZ_t = v(Z_t, t)dt$, 
to transport samples between a data distribution $\pi_0$ and a Gaussian noise distribution $\pi_1$. Sampling a clean image $X_0 \sim \pi_1$ and a Gaussian noise $X_1 \sim \pi_1$, the training objective is to make this velocity field $v$ 
match the direction of the straight-line path, $Z_t = tX_1 + (1 - t)X_0$, 
where $t \in [0, 1]$, achieved by minimizing the following least squares objective:
\begin{equation}
    \label{equ:rf_train_objective}
    \min_{v_\theta} \int_0^1 \mathbb{E} \left[ \left\| (X_1 - X_0) - v_\theta(Z_t, t) \right\|^2 \right] dt \quad
\end{equation}
The resulting model $v_\theta$ allows for backward sampling ($\pi_1 \to \pi_0$). 
In typical text-to-image generative tasks, 
this continuous ODE process is discretized, 
and the velocity field $v_\theta$ is 
additionally conditioned on a text prompt $P$, 
denoted as $v_\theta(Z_t, t, P)$. 
The image generation is achieved by 
solving the ODE backward from $t=1$ to $t=0$, 
% This process is discretized 
which is discretized 
using the Forward Euler method as follows: 
\begin{equation}
    \label{equ:rf_generation_process}
    Z_{t-\Delta t} = Z_t - v_\theta(Z_t, t, P)\Delta t
\end{equation}

\noindent\textbf{Inversion-Free Image Editing.}
In text-based image editing, flow models can be leveraged to construct a direct editing trajectory $Z^{edit}_t$ between a source image $X^{src}(t=1)$ and a target image $X^{tar}(t=0)$ described by the target prompt $P^{tar}$ \cite{kulikov2025flowedit}. 
Specifically, the source trajectory $Z^{src}_t$ is given by the linear interpolation, $Z^{src}_t = tX_1 + (1 - t)X^{src}$, where $X_1 \sim \pi_1$ is a Gaussian noise.
Thus, we can construct a dynamic target trajectory related to $Z^{src}_t$ and $Z^{edit}_t$: 
\begin{equation}
    \label{equ:target_path}
    Z^{tar}_t = Z^{src}_t + Z^{edit}_t - X^{src}
\end{equation}
Consequently, we obtained the editing velocity as follows:
\begin{equation}
    \label{equ:editing_velocity}
    % \begin{split}
    v_\theta^{edit}(t) = v_\theta(Z^{tar}_t, t, P^{tar}) - v_\theta(Z^{src}_t, t, P^{src})
    % \end{split}
\end{equation}
Similar to Eq.~\ref{equ:rf_generation_process}, we update the editing trajectory as follow:
\begin{equation}
    \label{equ:flowedit_editing_process}
        Z^{edit}_{t-\Delta t} = Z^{edit}_t - v_\theta^{edit}(t)\Delta t
\end{equation}

\subsection{Task Formulation}
\label{sec:task_definition}
\begin{figure*}[tb] 
    \centering
    \includegraphics[width=1.0\linewidth]{figs/pipeline2.2.pdf}
    \caption{\textbf{FlowDC Pipeline.} Given a source image, a source prompt and a target prompt, we first employ a LLM to decouple the complex prompt into a set of intermediate target prompts. Along the parallel editing trajectories guided by these prompts, the complex editing direction is decoupled into an orthogonal basis, as detailed in Sec.~\ref{sec:direction_decouple}. Then, as described in Sec.~\ref{sec:velocity_decay}, the editing velocity is decomposed and its components selectively decay during reconstrcution.}
    \label{fig:pipeline}
\end{figure*}

The inputs for our complex image editing task are a source image $X^{src}$, a source prompt $P^{src}$ and a complex target prompt $P^{tar}$. The complex target prompt comprises $n$ editing targets $\{e^{i}\}_{i=1}^{n}$. An editing target is defined as modifying a specific object or attribute with a specific editing type (e.g., change, addition, removal). For instance, the example in Figure~\ref{fig:pipeline} illustrates a complex prompt with editing targets such as: changing the color, modifying the shape, altering the cake's surface, adding a car and adding strawberries.

\subsection{FlowDC}

\subsubsection{Overview}
As illustrated in Figure~\ref{fig:pipeline}, we adopt the inversion-free flow to construct the editing trajectory from the source to the target. However, when directly conditioned on a complex prompt, the original editing velocity $v^{edit}(t)$ derived from Eq.~\ref{equ:editing_velocity} often produces suboptimal semantic alignment and imprecise editing behavior. 
Thus, our goal is to derive a more precise editing velocity $v^{edit'}(t)$, which enables the construction of a stable and accurate editing trajectory $Z^{edit}_t$, supporting faithful complex editing within a single round. To this end, we propose two techniques to improve velocity precision:
1) \textbf{Progressive Semantic Orthogonalization.} 
We first decouple the complex target prompt into a series of intermediate target prompts, which are used to construct multiple editing trajectories, similar to Eq.~\ref{equ:flowedit_editing_process}. The resulting set of editing directions is then decoupled into an orthogonal basis, each designed to isolate the semantic contribution of a specific editing target.
2)\textbf{Velocity Orthogonal Decay.} 
To reconstruct the final, precise velocity, the original editing velocity is projected onto a time-varying subspace, which is spanned by the orthogonal basis obtained in PSO (or alternatively, via displacement) at time $t$. 
During this reconstruction, the in-subspace component is primarily retained, while the orthogonal component strongly decays.

\subsubsection{Progressive Semantic Orthogonalization}
\label{sec:direction_decouple}

To derive the direction for each editing target $\{e^{i}\}_{i=1}^{n}$ (defined in Sec.~\ref{sec:task_definition}) along the editing trajectory $Z^{edit}_t$ and ensure alignment with the complex prompt $P^{tar}$, we propose the Progressive Semantic Orthogonalization (PSO) method.
First, we decouple the complex prompt into a series of intermediate target prompts. 
Guided by these prompts, we conduct parallel velocity generation to construct a set of parallel editing trajectories. 
Through progressive orthogonalization of these trajectory directions, we derive an orthogonal basis in the velocity subspace. 
Each basis vector isolates the semantic contribution of its corresponding editing target, guiding the preservation of complex editing semantics in the subsequent decay and reconstruction of the velocity.

\noindent\textbf{Complex Prompt Decoupling.}
We employ an LLM to decouple a complex prompt $P^{tar}$ (containing $n$ editing targets $\{e^i\}_{i=1}^n$) into an ordered sequence of $n$ intermediate target prompts, $\{P^{tar_i}\}_{i=1}^{n}$. 
Each intermediate prompt $P^{tar_i}$ is designed to cumulatively incorporate the first $i$ editing targets, $\{e^{j}\}_{j=1}^{i}$. The final prompt $P^{tar_n}$ is identical to the original complex prompt $P^{tar}$.

\noindent\textbf{Parallel Velocities Generation.}
Guided by the intermediate target prompts $\{P^{tar_i}\}_{i=1}^{n}$, our goal is to construct $n$ corresponding editing trajectories $\{Z^{edit_{i}}_t\}_{i=1}^n$ in parallel. These trajectories share a common source path, $Z_t^{src}$, defined by sampling a single Gaussian noise vector $X_1 \sim \pi_1$ and interpolating: $Z_t^{src}=tX_1 + (1 - t)X^{src}$. Consequently, we obtain a set of parallel target trajectories:
\begin{equation}
    \{Z^{tar_{i}}_t\}_{i=1}^n = \{Z_t^{src}+Z^{edit_{i}}_t - X^{src} \}_{i=1}^n
\end{equation}
To enable the subsequent construction and orthogonalization of the editing trajectories, we perform the parallel velocities generation (PVG) relying on the shared source trajectory $Z_t^{src}$ and the individual target trajectories $\{Z^{tar_{i}}_t\}_{i=1}^n$, as given by:
\begin{equation}
    \begin{split}
        \{v^{edit_{i}}(t)\}_{i=1}^n &= PVG(\{P^{tar_i}\}_{i=1}^n, t) \\
        &= \{v_\theta(Z^{tar_i}_t, t, P^{tar_i})- v^{src}(t)\}_{i=1}^n, \\
        v^{src}(t) &= v_\theta(Z^{src}_t, t, P^{src}), \\
    \end{split}
\end{equation}
For notational convenience, we denote $v^{edit_{i}}(t)$ as $v^i(t)$, $v^{edit}(t)$ as $v(t)$, $Z^{edit_{i}}_t$ as $Z^i_t$ and $Z^{edit}_t$ as $Z_t$. With these editing velocities computed, the discretized update rule from ODE solution for each editing trajectory is given by:
\begin{equation}
    \{Z^{i}_{t-\Delta t}\}_{i=1}^n = \{Z^i_t - v^i(t)\Delta t\}_{i=1}^n
\end{equation}

\noindent\textbf{Progressive Vectors Orthogonalization.}
To derive an orthogonal basis in the velocity subspace, where each basis vector isolates the novel semantic contribution of a single editing target, we propose the Progressive Vectors Orthogonalization (PVO). 
This process transforms a set of input ``editing vectors" $\{V^i(t)\}_{i=1}^{n}$ (\ie, editing velocities or displacement vectors at time $t$) into a mutually orthogonal set of sub-vectors $\{u_i(t)\}_{i=1}^{n}$. 
Specifically, each $u_i(t)$ is initialized as the input vector $V^i(t)$ and then iteratively refined by subtracting its projection onto each previously computed orthogonal vector $\{u_j(t)\}_{j=1}^{i-1}$, as detailed in Algorithm~\ref{pseudocode:vectors_orthogonalization}.

\begin{algorithm}[tb]
    \caption{Progressive Vectors Orthogonalization (PVO)}
    \label{pseudocode:vectors_orthogonalization}
\begin{algorithmic}
    \STATE {\bfseries Input:} 
    a set of editing vectors $\{V^i(t)\}_{i=1}^{n}$ 
    \STATE {\bfseries Output:} a set of orthogonal sub-vectors $\{u_i(t)\}_{i=1}^{n}$
    \FOR{$i=1$ {\bfseries to} $n$}
    \STATE $u_i(t) \leftarrow V^i(t)$
    \FOR{$j=1$ {\bfseries to} $i-1$}
    % \STATE $u_i(t) \leftarrow u_i(t) - \frac{\langle v_\theta^i(t), u_j(t) \rangle}{\|u_j(t)\|_2^2} u_j(t)$
    \STATE $u_i(t) \leftarrow u_i(t) - \frac{\langle u_i(t), u_j(t) \rangle}{\|u_j(t)\|_2^2} u_j(t)$
    \ENDFOR
    \ENDFOR
    \STATE {\bfseries Return:} $\{u_i(t)\}_{i=1}^{n}$
\end{algorithmic}
\end{algorithm}

\subsubsection{Velocity Orthogonal Decay}
\label{sec:velocity_decay}
To obtain a precise editing velocity $v'(t)$, we further propose Velocity Orthogonal Decay (VOD). 
1) It first decouples the original velocity $v(t)$ by projecting it onto the orthogonal basis $\{u_i(t)\}_{i=1}^{n}$ derived from PSO (Sec.~\ref{sec:direction_decouple}). 
2) Then during reconstruciton, the in-subspace component, which corresponds to the complex editing targets, is primarily retained, while the orthogonal component strongly decays to preserve the source consistency.
Combining this decay strategy with PSO ensures the reconstructed velocity $v'(t)$ focuses purely on the desired edits, simultaneously preserving source consistency and maintaining semantic alignment with the complex target prompt.

\noindent\textbf{Subspace Construction. }
In practice, we decouple this velocity onto a time-varying subspace $\mathbf{U}(t)$ as follows:
\begin{equation}
    \label{equ:subspace_definition1}
    \mathbf{U}(t) = 
        \begin{cases}
            PVO(\{d^i\}_{i=1}^n), & t \geq t_o \\
            \{d^n(t)\}, & t < t_o
        \end{cases}
\end{equation}
Here, $d^i(t)$ denotes the displacement of the $i$-th editing trajectory, $Z^i_t$, at time $t$. As we conduct the PVO process only in the early stage (for $t \geq t_o$), the displacement $d^n(t)$ used in the later stage (for $t < t_o$) already contains the semantic contributions of all editing targets. This two-stage approach is intentional, as conducting the PVO process at every timestep would require generating all $n$ parallel editing trajectories, which is computationally intensive.
\begin{algorithm}[tb]
    \caption{Overall Pipeline of FlowDC}
    \label{pseudocode:total_trajectory}
\begin{algorithmic}
    \STATE {\bfseries Input:} 
    source image $X^{src}$, 
    source prompt $P^{src}$, 
    complex target prompt $P^{tar}$, 
    time steps $t_1, t_g, t_o$
    \STATE {\bfseries Output:} edited image $Z_0$ 
    \STATE $\{P^{tar_i}\}_{i=1}^n \gets LLM(P^{src}, P^{tar})$
    \STATE $\{Z^i_{t_1}\}_{i=1}^n \gets \{X^{src}\}_{i=1}^n$
    \STATE $\{v^i_{g}\}_{i=1}^n \gets PVG( \{P^{tar_i}\}_{i=1}^n, t_g)$
    \FOR{$t: t_1 \rightarrow 0$}
        \FOR{$i: 1 \rightarrow n$}
        \STATE $v^i(t) \gets PVG(P^{tar_i}, t)$ \COMMENT{original}
        \STATE $d^i \gets Z^i_t - X^{src}$
        \STATE $\mathbf{U}_i(t) \gets d^i$ 
        \IF{$t = t_1$} %\tikzmark{start-pso}
            \STATE $\mathbf{U}_i(t) \gets PVO(\{v^i_g\}_{j=1}^i)$
        \ELSIF{$t \geq t_o$}
            \STATE $\mathbf{U}_i(t) \gets PVO(\{d^j\}_{j=1}^i)$  
        % \ELSE %\tikzmark{anchor-pso}
        %     \STATE $\mathbf{U}_i(t) \gets d^i$  
        \ENDIF %\tikzmark{end-pso}
        \STATE $v_{sub}^i(t) \gets Proj(v^i(t), \mathbf{U}_i(t))$
        \STATE $v_{orth}^i(t) \gets v^i(t) - v_{sub}^i(t)$
        \STATE $v^{i}{'}(t) \gets v_{sub}^i(t) + \lambda_{orth}(t)v_{orth}^i(t)$ \COMMENT{precise} 
        \STATE $Z^i_t \gets Z^i_t - v^{i}{'}(t)\Delta t$
        \ENDFOR
    \ENDFOR
    \STATE $Z_0 \gets Z^n_0$
\end{algorithmic}
\end{algorithm}

However, at the initial time $t_1$, the displacements $\{d^i(t_1)\}_{i=1}^{n}$ are zero, which would cause the subspace $\mathbf{U}(t_1)$ to collapse. Concurrently, the velocity $v(t_1)$ is highly noisy due to the high interpolation weight of Gaussian noise. We address this by computing a set of reference velocities, $\{v^i(t_g)\}_{i=1}^{n}$, at a designated guidance time $t_g$. These velocities are then used as the input to the PVO process at $t_1$, in place of the zero-valued displacements $\{d^{i}(t_1)\}_{i=1}^{n}$. We rewrite the subspace $\mathbf{U}(t)$ as follows:
\begin{equation}
    \label{equ:subspace_definition2}
    \mathbf{U}(t) = 
        \begin{cases}
            PVO(\{v^i(t_g)\}_{i=1}^n), & t = t_1 \\
            PVO(\{d^i\}_{i=1}^n), & t_o \leq t < t_1 \\
            \{d^n(t)\}, & t < t_o
        \end{cases}
\end{equation}

\noindent\textbf{Velocity Decay and Reconstruction. }
As $\mathbf{U}(t)$ represents an orthogonal basis, the in-subspace component $v_{sub}(t)$ of the editing velocity $v(t)$ is defined:
\begin{equation}
    \label{equ:projection}
    \begin{split}
        v_{sub}(t) &= \text{Proj}(v(t), {\mathbf{U}(t)}) \\
        &= \sum_{u_j(t) \in \mathbf{U}(t)} \frac{\langle v(t), u_j(t) \rangle}{\|u_j(t)\|_2^2} u_j(t)
    \end{split}
\end{equation}
Consequently, the orthogonal component, $v_{orth}(t)$, is the remaining part of the original velocity, $v_{orth}(t) = v(t) - v_{sub}(t)$.
After decoupling the editing velocity $v(t)$, we reconstruct the modified velocity $v'(t)$ via selectively decay:
\begin{equation}
    \label{equ:velocity_reconstuct}
    v'(t) = 
            \lambda_{sub}(t)v_{sub}(t)+\lambda_{orth}(t)v_{orth}(t)
\end{equation}
To preserve the editing semantics, we set the in-subspace coefficient $\lambda_{sub}(t) = 1$ for all $t$. The orthogonal coefficient $\lambda_{orth}(t)$ is defined by a piecewise linear decay:
\begin{equation}
    \label{equ:lambda_orth}
    \lambda_{orth}(t) = 
    \begin{cases}
        \lambda_d + \frac{(\lambda_1-\lambda_d)(t-t_d)}{t_1-t_d}, & t \geq t_d \\
        1, & t < t_d
    \end{cases}
\end{equation}
Here, $\lambda_1$ and $\lambda_d$ are hyperparameters controlling the decay strength.
Finally, this modified editing velocity $v'(t)$ is used to update the main editing trajectory $Z_t$, (i.e., $Z^n_t$):
\begin{equation}
    \label{equ:final_update_rule}
     Z_{t-\Delta t} = Z_t - v'(t)\Delta t
\end{equation}
The complete process is summarized in Algorithm~\ref{pseudocode:total_trajectory}.

\subsection{Complex-PIE-Bench}
\label{sec:complex_pie_bench}
Existing benchmarks cannot adequately evaluate the complex image editing task defined in Sec.~\ref{sec:task_definition}. 
Complex-Edit~\cite{yang2025textttcomplexeditcotlikeinstructiongeneration} benchmark 
takes instructions as complex editing inputs, 
which is not aligned with our source-target prompt pair setting.
PIE-Bench++~\cite{huang2024paralleledits} benchmark 
derived from PIE-Bench~\cite{ju2023direct} 
evaluates the performance of complex image editing. 
However, only $19\%$ contains three or more editing targets.
Therefore, to facilitate a more rigorous evaluation of complex image editing, we propose a new benchmark that extends PIE-Bench~\cite{ju2023direct}. 
% \subsubsection{Benchmark Construction}

\noindent\textbf{Benchmark Construction.}
The original PIE-Bench provides samples with only a single target prompt $P^{tar_1}$. 
To construct our benchmark, \textbf{Complex-PIE-Bench}, we employ the pre-trained multi-modal model (doubao-seed-1.6) to extend to a set of four editing target prompts. 
Therefore, Complex-PIE-Bench contains four editing targets and four intermediate target prompts $\{P^{tar_i}\}_{i=1}^{4}$ per sample, enabling  both single-round and multi-round image editing\footnote{More details are placed in appendix.\label{appendix}}.

\section{Experiments}
\label{sec:experiments}

\noindent\textbf{Implementation details.}
We use FLUX.1 dev ~\cite{labs2025flux1kontextflowmatching}
as the base flow model in our experiments. 
We set the same guidance scales as FlowEdit ~\cite{kulikov2025flowedit} of $1.5$ and $5.5$ for the source and target conditioning, respectively. 
For time steps hyperparameters, we use  
$T=28$ steps, with $t_{1}=27/28, t_{g}=22/28, t_{o}=27/28$ in Eq.~\ref{equ:subspace_definition2}. 
For the factor $\lambda$ in Eq.~\ref{equ:lambda_orth}, 
we set it as follows: 
$\lambda_{1}=0.1, \lambda_{d}=0.64, t_{d}=20/28$.

\noindent\textbf{Evaluation Benchmarks.}
We evaluate complex editing performance on two benchmarks: PIE-Bench++~\cite{huang2024paralleledits} and our new Complex-PIE-Bench. PIE-Bench++ offers a varied distribution of editing complexity (samples with 1, 2, or 3+ targets), whereas Complex-PIE-Bench consists exclusively of samples with exactly four editing targets. Each benchmark contains 700 samples covering the 10 editing categories defined in PIE-Bench~\cite{ju2023direct}.

\noindent\textbf{Baselines.}
We compare our method to several competing text-based image editing methods for flow models. For single-round editing, we compare the inversion-free method FlowEdit~\cite{kulikov2025flowedit}, as well as the inversion-based methods RF-Inversion~\cite{rf_inversion}, RF-Edit~\cite{rf-solver}, SDEdit~\cite{mengsdedit} and vanilla ODE inversion. For multi-round editing, we compare Multi-Turn~\cite{zhou2025multi}. 
% We also compare to instruction-guided, training-based methods such as OmniGen~\cite{xiao2025omnigen} and Instruct-CLIP~\cite{chen2025instruct}, evaluating them using their officially provided pretrained weights.
Furthermore, we compare to ParallelEdit~\cite{huang2024paralleledits}, a single-round editing method that manipulates the attention layers of diffusion models. However, this approach requires complex prompt engineering, and its official implementation can only handle a subset of PIE-Bench++. Consequently, we limit our comparison in the main text to a qualitative evaluation of their official implementation and provide partial quantitative results in Appendix\footref{appendix}.

\subsection{Qualitative evaluation}
\begin{figure*}[tb] 
    \centering
    \includegraphics[width=1.0\textwidth]{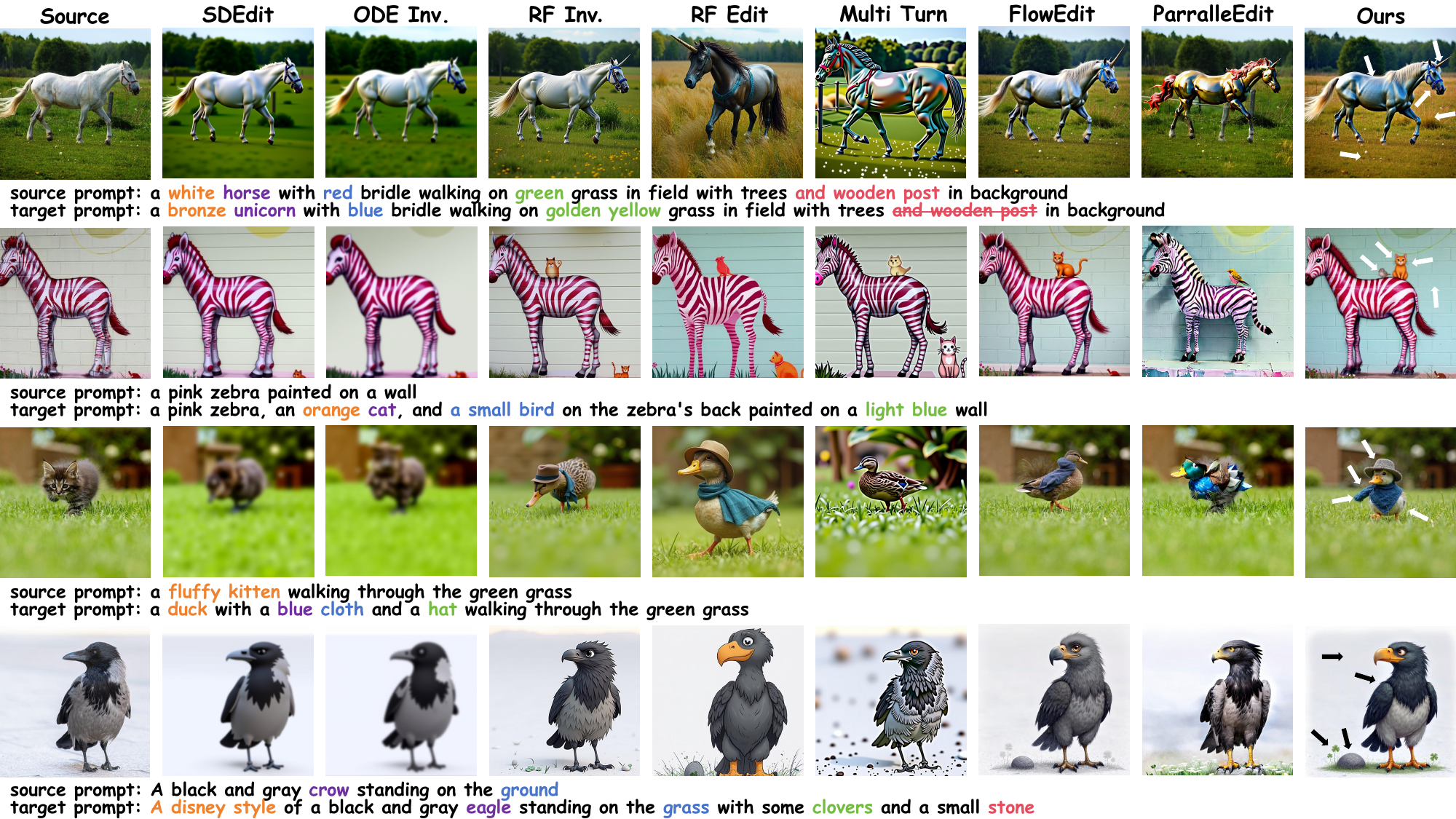}
    \caption{\textbf{Qualitative Comparison of complex editing.} Editing targets are indicated by differently colored text and arrows.}
    \vspace{-1em}\label{fig:quality_result}
\end{figure*}

As shown in Fig.~\ref{fig:quality_result}, we observe the following: 
1) SDEdit, ODE-Inversion, RF inversion and ParallelEdit exhibit poor semantic alignment with the complex editing targets. For example, in the first row, RF Inversion fails to update the horse's color, while ParallelEdit neglects to change the color of the grass.
2) RF Edit and Multi-turn suffer from poor source consistency. For instance, in the third row, RF Edit alters the duck's orientation and background, while Multi-turn incorrectly modifies the background.
3) FlowEdit~\cite{kulikov2025flowedit} demonstrate a better trade-off than other baselines. However, it still fails to successfully achieve the complex editing targets while fully preserving non-target regions. In the second row, for example, it fails to add a small bird. 
4) In contrast, our method achieves a good balance between semantic alignment and source consistency.

\subsection{Quantitative evaluation}

\noindent\textbf{Setting.} 
We assess two key aspects: complex semantic alignment using CLIP-T~\cite{clip}, and source consistency using CLIP-I, DINO~\cite{zhang2022dino}, and LPIPS~\cite{zhang2018unreasonable}.

\noindent\textbf{Results.} 
As shown in Table~\ref{tab:quantitative_result}, we can observe: 1) Our method exhibits the best source consistency, achieving the top scores across all three related metrics: CLIP-I, DINO, and LPIPS. 2) RF-Edit excels in semantic metrics (CLIP-T) but performs poorly on source consistency (DINO, LPIPS), indicating it over-edits the image. 3) Our method also achieves strong performance in semantic alignment, securing the second-best score on the CLIP-T metric. 4) Consequently, our method achieves an excellent balance between source consistency and semantic alignment.

\begin{table*}[t]
    \centering
    \renewcommand{\arraystretch}{0.9}
    \setlength{\aboverulesep}{2pt}
    \setlength{\belowrulesep}{2pt}
    \caption{\textbf{Quantitative comparison on Complex-PIE-Bench and PIE-Bench++.} 
    Metrics of CLIP-T, CLIP-I, DINO and LPIPS are reported.
    For all metrics except LPIPS, higher is better ($\uparrow$). Best results are in \textbf{bold} and second best are \underline{underlined}.}
    \label{tab:quantitative_result}
    \resizebox{1.8\columnwidth}{!}{%
    \def\arraystretch{0.9}
    \begin{tabular}{llcccc}
        \hline
        \hline
        \textbf{Dataset} & \textbf{Method} & \textbf{CLIP-T (\%)} $\uparrow$ & \textbf{CLIP-I (\%)} $\uparrow$ & \textbf{DINO (\%)} $\uparrow$ & \textbf{LPIPS (\%)} $\downarrow$ \\
        \midrule
        \multirow{7}{*}{Complex-PIE-Bench} 
         & ODE-Inv & 22.78 & 81.89 & 64.99 & 51.54 \\
         & SDEdit~\cite{mengsdedit} & 23.59 & 85.17 & 70.55 & 44.10 \\
         & MultiTurn~\cite{zhou2025multi} & 25.03 & 81.11 & 61.51 & 51.30 \\
         & RF-Inv~\cite{rf_inversion} & 26.59 & 85.79 & 68.31 & 37.94 \\
         & FlowEdit~\cite{kulikov2025flowedit} & 26.91 & \underline{87.63} & \underline{70.84} & \underline{23.86} \\
         & RF-Edit~\cite{rf-solver} & \textbf{28.79} & 78.47 & 47.62 & 50.80 \\
        \cmidrule(l){2-6}
        & Ours & \underline{27.69} & \textbf{87.72} & \textbf{71.69} & \textbf{23.72} \\
        \midrule
        \multirow{7}{*}{PIE-Bench++} 
         & ODE-Inv & 21.75 & 81.22 & 62.94 & 51.78 \\
         & SDEdit~\cite{mengsdedit} & 22.16 & 84.93 & 69.67 & 44.23 \\
         & MultiTurn~\cite{zhou2025multi} & 23.87 & 83.85 & 63.72 & 44.34 \\
         & RF-Inv~\cite{rf_inversion} & 24.59 & 85.82 & 67.69 & 37.51 \\
         & FlowEdit~\cite{kulikov2025flowedit} & 25.12 & \underline{88.27} & \underline{70.26} & \underline{22.44} \\
         & RF-Edit~\cite{rf-solver} & \textbf{26.88} & 78.04 & 45.35 & 49.12 \\
        \cmidrule(l){2-6}
        & Ours & \underline{25.13} & \textbf{88.76} & \textbf{72.33} & \textbf{22.09} \\
        \hline
        \hline
    \end{tabular}
    }
    \vspace{-1.0em}
\end{table*}

\begin{figure}[t] 
    \centering
    \includegraphics[width=1.0\linewidth]{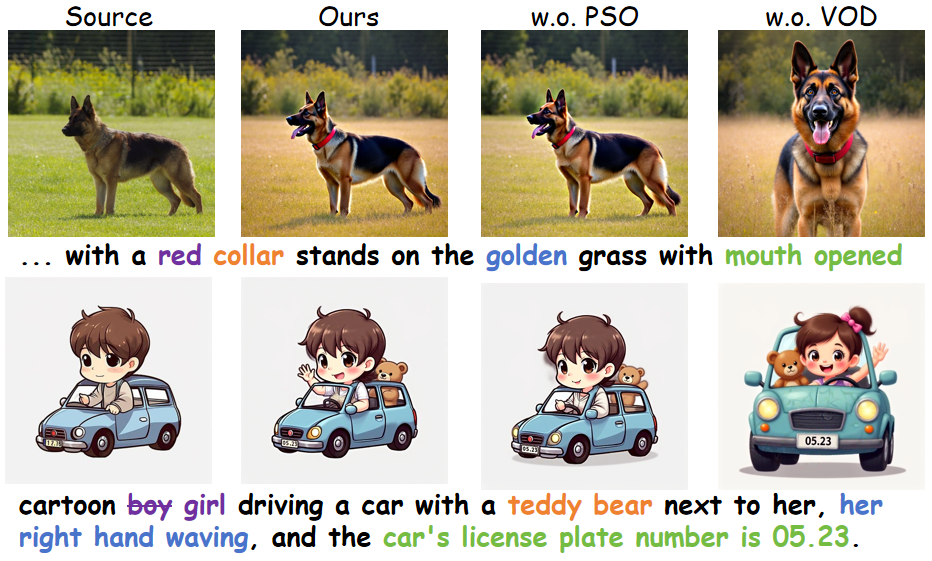}
    \caption{Ablation study of design components.}
    \vspace{-1.5em}\label{fig:ablation_example}
\end{figure}

\begin{table}[htbp]
    \centering
    \caption{Ablation Study on Complex-PIE-Bench.}
    \label{tab:quantitative_ablation}
    \resizebox{1\columnwidth}{!}{%
    % \scriptsize
    \setlength{\tabcolsep}{3pt}
    \begin{tabular}{lcccc}
        \hline
        % \hline
        \textbf{Method} & \textbf{CLIP-T (\%)} $\uparrow$ & \textbf{CLIP-I (\%)} $\uparrow$ & \textbf{DINO (\%)} $\uparrow$ & \textbf{LPIPS (\%)} $\downarrow$ \\
        \toprule
         \textbf{Ours}           & \underline{27.69} & \underline{87.72} & \textbf{71.69} & \textbf{23.72} \\
         \textbf{w.o. PSO} & 27.30 & \textbf{87.76} & \underline{71.60} & \underline{24.19} \\
         \textbf{w.o. VOD} & \textbf{29.23} & 79.63 & 47.58 & 49.89 \\
        
        % \hline
        \hline
    \end{tabular}
    }
    \vspace{-0.5em}
\end{table}

\subsection{Ablation Study}

We conduct ablation studies to validate our two core designs: Progressive Semantic Orthogonalization (PSO) and Velocity Orthogonal Decay (VOD). 
As illustrated both qualitatively in Fig.~\ref{fig:ablation_example} and quantitatively in Table~\ref{tab:quantitative_ablation}, these designs serve critical roles. 
PSO proves essential for semantic alignment with the complex editing targets. 
Qualitatively, its absence (Fig.~\ref{fig:ablation_example}) results in a failure to satisfy complex targets, such as modifying the grass's color to golden or making the girl wave. 
This observation is quantitatively substantiated in Table~\ref{tab:quantitative_ablation}, where ablating PSO causes a distinct drop in the semantic alignment metric, CLIP-T.
VOD is crucial for preserving source consistency. 
Qualitatively, its removal (Fig.~\ref{fig:ablation_example}) leads to a significant loss of structural integrity, disrupting elements like the dog's background and the car's orientation. 
This structural collapse is numerically reflected in Table~\ref{tab:quantitative_ablation}, where ablating VOD causes a severe degradation in all source consistency metrics.

\begin{figure}[t]
    \centering
    \includegraphics[width=1.0\linewidth]{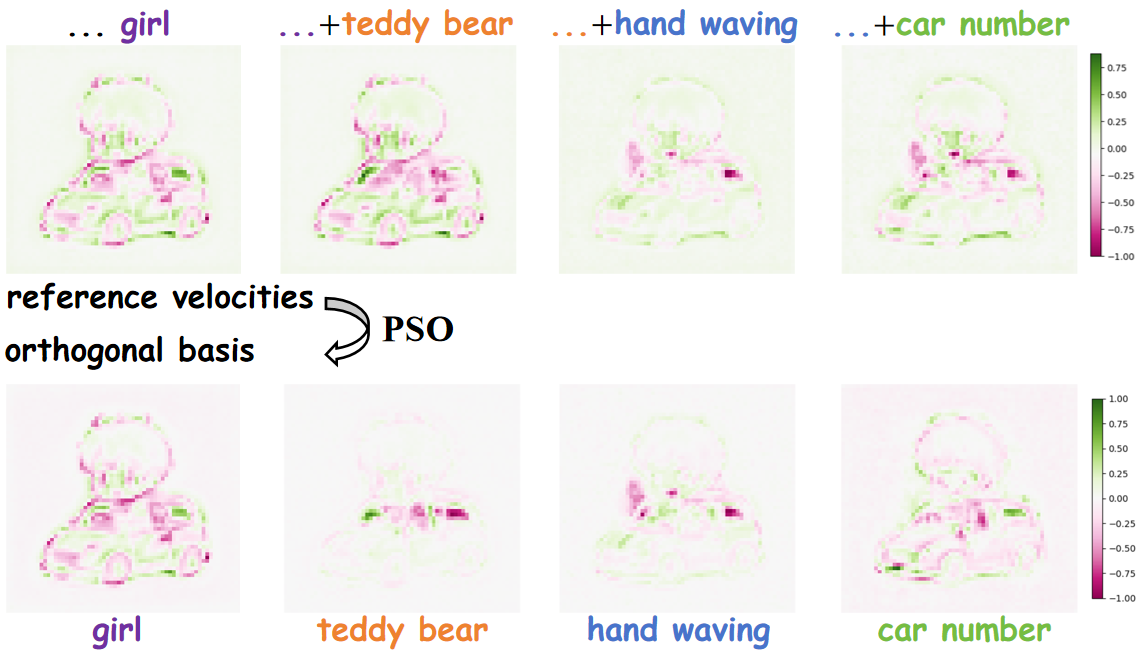}
    \caption{The orthogonal basis's heatmaps in PSO.
    }
    \label{fig:pso_ablation.png}
    \vspace{-1.5em}
\end{figure}

\noindent \textbf{PSO Analysis.} We visualize the orthogonal basis spanned by the reference velocities at the start time in PSO (Fig.~\ref{fig:pso_ablation.png}). We observe that each basis vector successfully isolates the semantic contribution of its corresponding editing target. For instance, the heatmap corresponding to the second basis vector clearly highlights the teddy bear area. This effective semantic decomposition is crucial for our method's ability to handle complex edits.

\begin{figure}[t]
    \centering
    \includegraphics[width=1.0\linewidth]{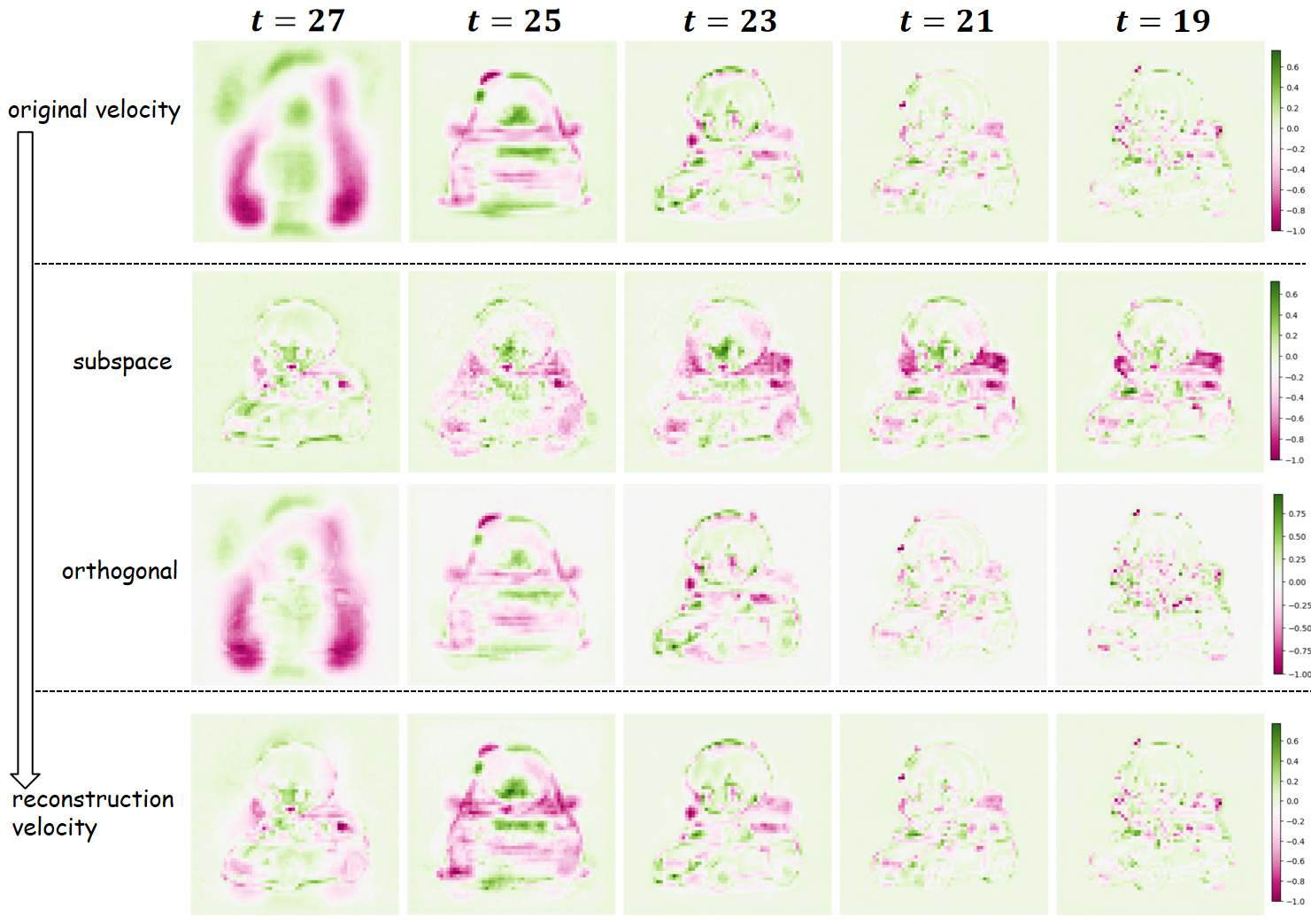}
    \caption{The velocities's heatmaps in VOD.}
    \label{fig:vod_ablation}
    \vspace{-1.5em}
\end{figure}

\noindent \textbf{VOD Analysis.} 
As shown in Fig.~\ref{fig:vod_ablation}, the heatmap visualization reveals that the orthogonal velocity component, $v_{orth}(t)$, can be detrimental to preserving the source structure. 
For instance, the heatmap corresponding to the orthogonal velocity disrupts the car's orientation, particularly during the initial timesteps. 
Therefore, VOD is applied to mitigate this issue, enabling the reconstruction velocity $v'(t)$ to guide a more precise edit than the original velocity $v(t)$.

\section{Conclusion}
In this paper, we analyzed the challenges of complex text-based image editing and the limitations of existing methods. To address these problems, we introduced FlowDC, a novel flow decoupling and decay method. For a more comprehensive evaluation of this task, we also constructed a new complex editing dataset. Extensive comparative and ablation experiments validate the effectiveness of our approach. 
Moving forward, we are going to 1) extend our method into other modalities, e.g., video editing; 2) adapt our method to support multi-modal guidance, such as combining text prompts with reference images or sketches.
\clearpage
{
    \small
    \bibliographystyle{ieeenat_fullname}
    \bibliography{main}
}
\clearpage
% \clearpage
% \setcounter{page}{1}

\maketitlesupplementary

\appendix

This supplementary document is organized as follows:
\pagestyle{empty}  % no page number for the second and the later pages
\thispagestyle{empty} % no page number for the first page
\begin{itemize}

    \item[$\bullet$]  In Sec.~\ref{sec:dataset}, we show more details about the construction and containment of Complex-PIE-Bench 
    \item[$\bullet$]  In Sec.~\ref{sec:additional_setting}, we describe the implementation details for both competing methods and our method, outlining how the quantitative and qualitative results were obtained. 
    \item[$\bullet$]  In Sec.~\ref{sec:additional_qualitative}, we present additional qualitative results.
    \item[$\bullet$]  In Sec.~\ref{sec:additional_quantitative}, we provide additional quantitative comparisons with ParallelEdit and analyze the performance trade-offs of our method and baselines under different configurations.
    \item[$\bullet$]  In Sec.~\ref{sec:trajectory_analysis}, we analyze the editing trajectory, which demonstrates how VOD maintains source consistency.
    \item[$\bullet$]  In Sec.~\ref{sec:flowdc_param}, we provide more ablation results and explain the selection of hyperparameters for our method
    \item[$\bullet$]  In Sec.~\ref{sec:limitation}, we discuss our work’s limitations.
    \item[$\bullet$]  In Sec.~\ref{sec:full_algorithm}, we provide the full algorithm of our method.
    \item[$\bullet$]  In Sec.~\ref{sec:mathematical_pso}, we provide the mathematical justification of PSO design.

\end{itemize}

\section{Datasets}
\label{sec:dataset}
Our work builds upon the original PIE-Bench, which provides samples with a single target prompt across 10 editing categories, outlined as follows:
0) Random;
1) Change object;
2) Add object;
3) Delete object;
4) Change attribute content;
5) Change attribute pose;
6) Change attribute color;
7) Change attribute material;
8) Change background;
9) Change style.

To construct our benchmark, Complex-PIE-Bench, we extend each initial sample from a single prompt to a set of four editing targets and prompts. We employed the pre-trained multi-modal model (doubao-seed-1.6) for this expansion. The model generates new pairs conditioned on the initial pair and the source image.

\begin{figure}[h]
    \centering
    \includegraphics[width=0.85\linewidth]{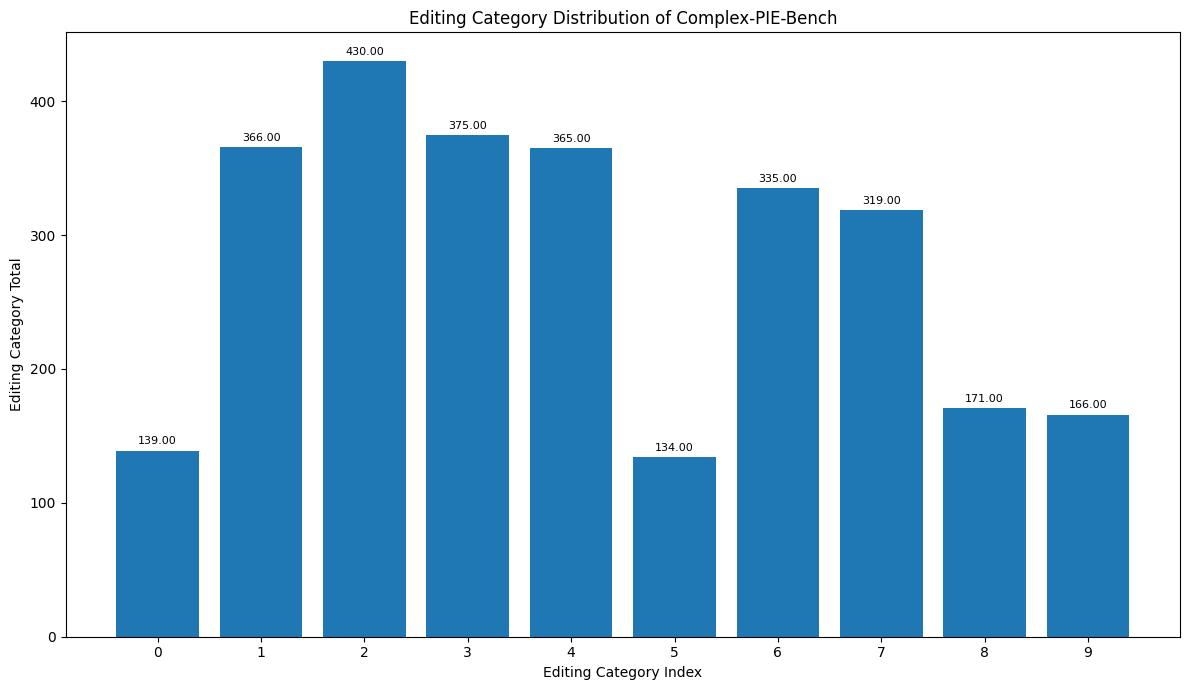}
    \vspace{-0.5em}
    \caption{Editing Category Distribution of Complex-PIE-Bench}
    \label{fig:editing_distribution}
\end{figure}

The generation process adheres to two main principles: 1) Editing targets must not be conflicting (e.g., first transforming an object's color and then deleting that same object). 2) Global editing categories (i.e., background modification, style transfer) and the object deletion category are not repeated within the same sample. The resulting editing category distribution for Complex-PIE-Bench is shown in Fig.~\ref{fig:editing_distribution}.

\section{Additional Experiments}
\label{sec:additional_experiments}

\subsection{Additional details on the experiment settings}
\label{sec:additional_setting}
In Sec.~\ref{sec:experiments}, we compare our method against seven competing approaches: 1) ODE Inversion, 2) SDEdit~\cite{mengsdedit}, 3) RF Inversion~\cite{rf_inversion}, 4) RF Edit~\cite{rf-solver}, 5) Multi Turn~\cite{zhou2025multi}, 6) FlowEdit~\cite{kulikov2025flowedit}, and 7) ParallelEdit~\cite{huang2024paralleledits}.
For each baseline (with the exception of ParallelEdit), we explored multiple sets of hyperparameter configurations on Complex-PIE-Bench. 
The hyperparameters adopted for the main experiments in Sec.~\ref{sec:experiments} are highlighted in \textbf{bold}. 

For ODE-Inv, SDEdit, and FlowEdit, following the setting in FlowEdit~\cite{kulikov2025flowedit}, we set total steps to 28. As detailed in Table~\ref{tab:hyperpara_flowedit}, the CFG scales for the source and target were set to 1.5 and 5.5, respectively, while the start timestep $t_1$ was varied.
\begin{table}[h]
    \caption{Hyperparameters for ODE-Inv, SDEdit, and FlowEdit.}
    % \vspace{-0.4em}
    \label{tab:hyperpara_flowedit}
    \centering
    \resizebox{0.5\columnwidth}{!}{%
    \begin{tabular}{lcccc}
        \hline
         Method & steps $T$ & source CFG & target CFG & start step $t_1$ \\
        \midrule
        ODE-Inv  & 28 & 1.5 & 5.5 & 20/28, \textbf{24/28} \\
        SDEdit   & 28 & -   & 5.5 & \textbf{21/28}, 24/28 \\
        FlowEdit & 28 & 1.5 & 5.5 & 23/28, \textbf{24/28}, 25/28 \\
        \hline
    \end{tabular}
    }
    % \vspace{-0.4em}
\end{table}

Regarding RF Inversion, we adhered to the official implementation settings: the control strength $\eta$ was fixed at $0.9$ and the starting timestep $s$ at $0$, with the stopping timestep $\tau$ being the variable parameter (see Table~\ref{tab:hyperpara_rf_inversion}).
\begin{table}[h]
    \caption{Hyperparameters for RF Inversion.}
    % \vspace{-0.4em}
    \label{tab:hyperpara_rf_inversion}
    \centering
    \resizebox{0.35\columnwidth}{!}{%
    \begin{tabular}{lcc}
        \hline
         Method & strength $\eta$ & stopping step $\tau$ \\
        \midrule
        RF Inversion  & 0.9 & 5/28, \textbf{7/28} \\
        \hline
    \end{tabular}
    }
    % \vspace{-0.4em}
\end{table}

For RF Edit, we adopted the standard guidance scale of $2$ and $30$ total steps, varying injection steps as shown in Table~\ref{tab:hyperpara_rf_edit}.
\begin{table}[h]
    \caption{Hyperparameters for RF Edit.}
    % \vspace{-0.4em}
    \label{tab:hyperpara_rf_edit}
    \centering
    \resizebox{0.3\columnwidth}{!}{%
    \begin{tabular}{lcc}
        \hline
         Method & guidance & injection steps \\
        \midrule
        RF Edit  & 2 & \textbf{2}, 4 \\
        \hline
    \end{tabular}
    }
    % \vspace{-0.4em}
\end{table}

For Multi Turn, consistent with the official codebase, we set the total time steps $T$ to $15$ and experimented with different guidance scales, as detailed in Table~\ref{tab:hyperpara_multi_turn}.
\begin{table}[h]
    \caption{Hyperparameters for Multi Turn.}
    % \vspace{-0.4em}
    \label{tab:hyperpara_multi_turn}
    \centering
    \resizebox{0.3\columnwidth}{!}{%
    \begin{tabular}{lcc}
        \hline
         Method & steps $T$ & guidance \\
        \midrule
        Multi Turn  & 15 & 2.5, \textbf{3.5} \\
        \hline
    \end{tabular}
    }
    % \vspace{-0.4em}
\end{table}

For ParallelEdit, we employed the official implementation built upon a diffusion model (LCMDreamshaperv7).

Finally, for our proposed method, we aligned our settings with FlowEdit~\cite{kulikov2025flowedit} by applying source and target guidance scales of 1.5 and 5.5, respectively. We set the total time steps to $T=28$, with key time steps configured as $t_{1}=27/28$, $t_{g}=22/28$, and $t_{o}=27/28$. The decay factor $\lambda(t)$ (Eq.~\ref{equ:lambda_orth}) was varied as presented in Table~\ref{tab:hyperpara_ours}.

\begin{table}[h]
    \caption{Hyperparameters for Ours.}
    % \vspace{-0.4em}
    \label{tab:hyperpara_ours}
    \centering
    \resizebox{0.5\columnwidth}{!}{%
    \begin{tabular}{lc}
        \hline
         Method & Decay Strength $(\lambda_1, \lambda_d, t_d)$ \\ 
        \midrule
        Ours  & (0.1, 1.0, 11/28), \textbf{(0.1, 0.64, 20/28)},  (0.3, 1.0, 20/28) \\
        \hline
    \end{tabular}
    }
    % \vspace{-0.4em}
\end{table}

\subsection{Additional Qualitative Results}
\label{sec:additional_qualitative}
\begin{figure*}[tb] 
    \centering
    \includegraphics[width=1.0\textwidth]{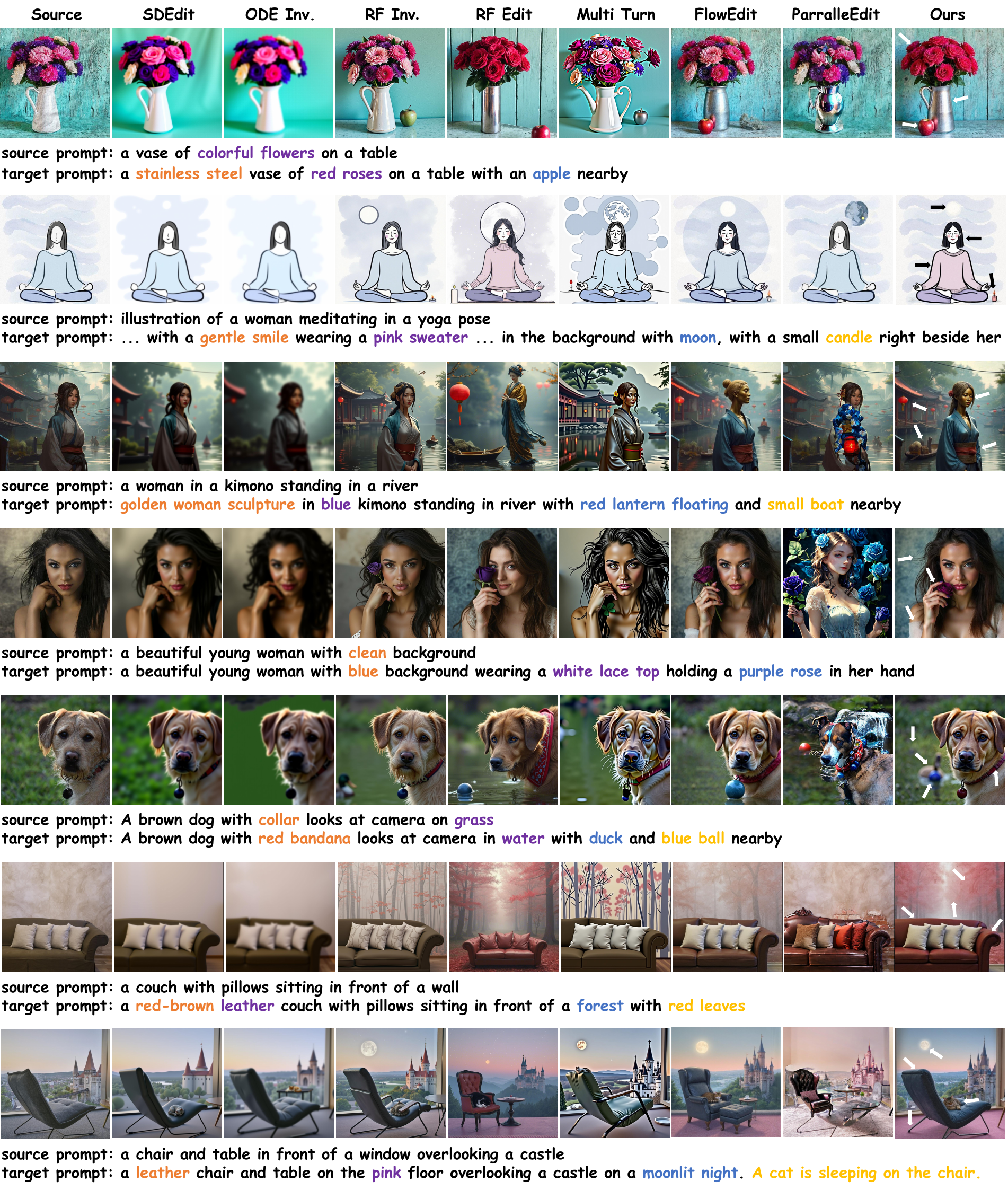}
    \caption{\textbf{Additional qualitative Comparison of complex editing.} Editing targets are indicated by differently colored text and arrows.}
    % \vspace{-1em}
    \label{fig:additional_qualitative_results}
\end{figure*}

Fig.~\ref{fig:additional_qualitative_results} shows additional qualitative comparisons between our method and other baselines.

\subsection{Additional Quantitative Results}
\label{sec:additional_quantitative}

\noindent\textbf{Comparison with ParallelEdit.}
As the official implementation of ParallelEdit is compatible with only a subset of PIE-Bench++, we limit our comparison to this specific subset, which contains $508$ samples of total $700$ samples. 
The decay strength parameters of our method $(\lambda_1, \lambda_d, t_d)$ are set to $(0.2, 1.0, 22/28)$. 
As shown in Table~\ref{tab:quantitative_result_paralleledit}, our method achieves slightly superior semantic alignment and comparable source consistency to ParallelEdit. Notably, while ParallelEdit relies on complex prompt engineering to obtain these results, our approach is more flexible and user-friendly, requiring only simple inputs.
\begin{table}[h]
    \centering
    % \renewcommand{\arraystretch}{0.9}
    % \setlength{\aboverulesep}{2pt}
    % \setlength{\belowrulesep}{2pt}
    % \vspace{-0.5em}
    \caption{Additional quantitative comparison with ParallelEdit.}
    % \vspace{-0.5em}
    \label{tab:quantitative_result_paralleledit} 
    \resizebox{0.6\columnwidth}{!}{%
    \begin{tabular}{lcccc}
        \hline
        % \hline
        \textbf{Method} & \textbf{CLIP-T (\%)} $\uparrow$ & \textbf{CLIP-I (\%)} $\uparrow$ & \textbf{DINO (\%)} $\uparrow$ & \textbf{LPIPS (\%)} $\downarrow$ \\
        \midrule
         ParallelEdit~\cite{huang2024paralleledits} & 26.47 & 84.22 & \textbf{59.88} & 32.42 \\
        % \midrule
        Ours & \textbf{26.51} & \textbf{84.47} & 58.84 & \textbf{31.46} \\
        
        % \hline
        \hline
    \end{tabular}
    }
    % \vspace{-0.5em}
\end{table}

\begin{figure}[h]
    \centering
    \includegraphics[width=0.75\linewidth]{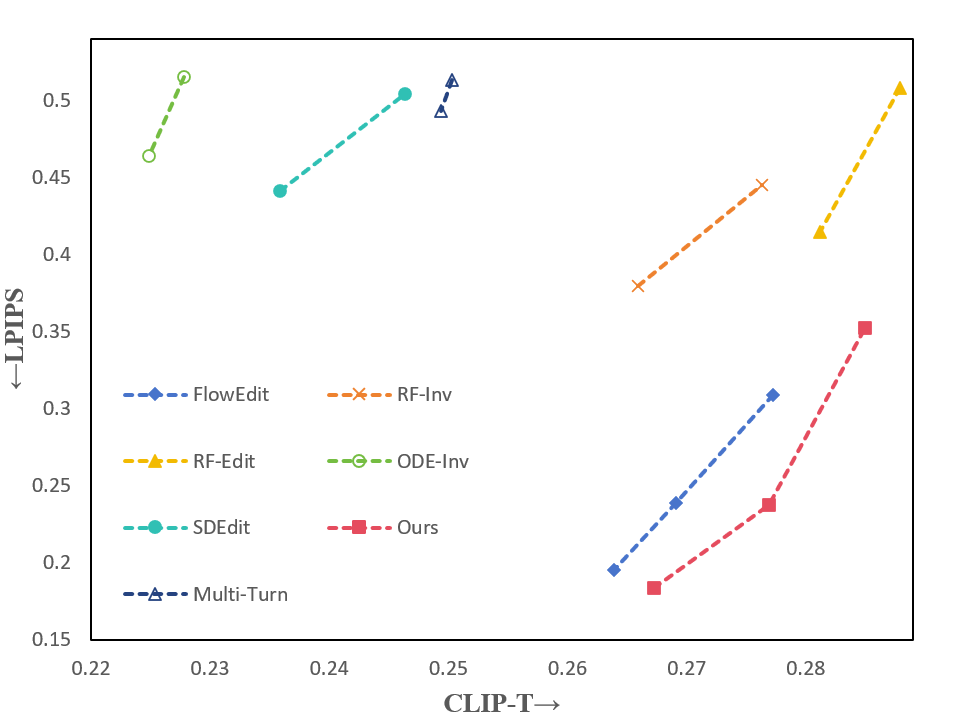}
    \caption{Additional quantitative comparisons.}
    \label{fig:quality_result2}
\end{figure}

\noindent\textbf{Additional Results.} 
While Table~\ref{tab:quantitative_result} presents our quantitative comparison using fixed hyperparameters, we conduct an additional quantitative comparison on Complex-PIE-Bench across a range of hyperparameter settings (detailed in Sec.~\ref{sec:additional_setting}), which allows us to intuitively map the performance trade-off for each method. 
We employ LPIPS to measure source consistency (lower is better) and CLIP-T to assess semantic alignment (higher is better).
The Fig.~\ref{fig:quality_result2} clearly demonstrate that FlowDC achieves a superior and more consistent balance across the performance envelope. In contrast, competing methods typically reveal a distinct compromise: they either rigidly preserve the original image structure, resulting in insufficient editing intensity, or they heavily modify the image, significantly compromising its source consistency.

\subsection{User Study}
\label{sec:user_study}
\begin{figure}[h]
    \centering
    \includegraphics[width=0.8\linewidth]{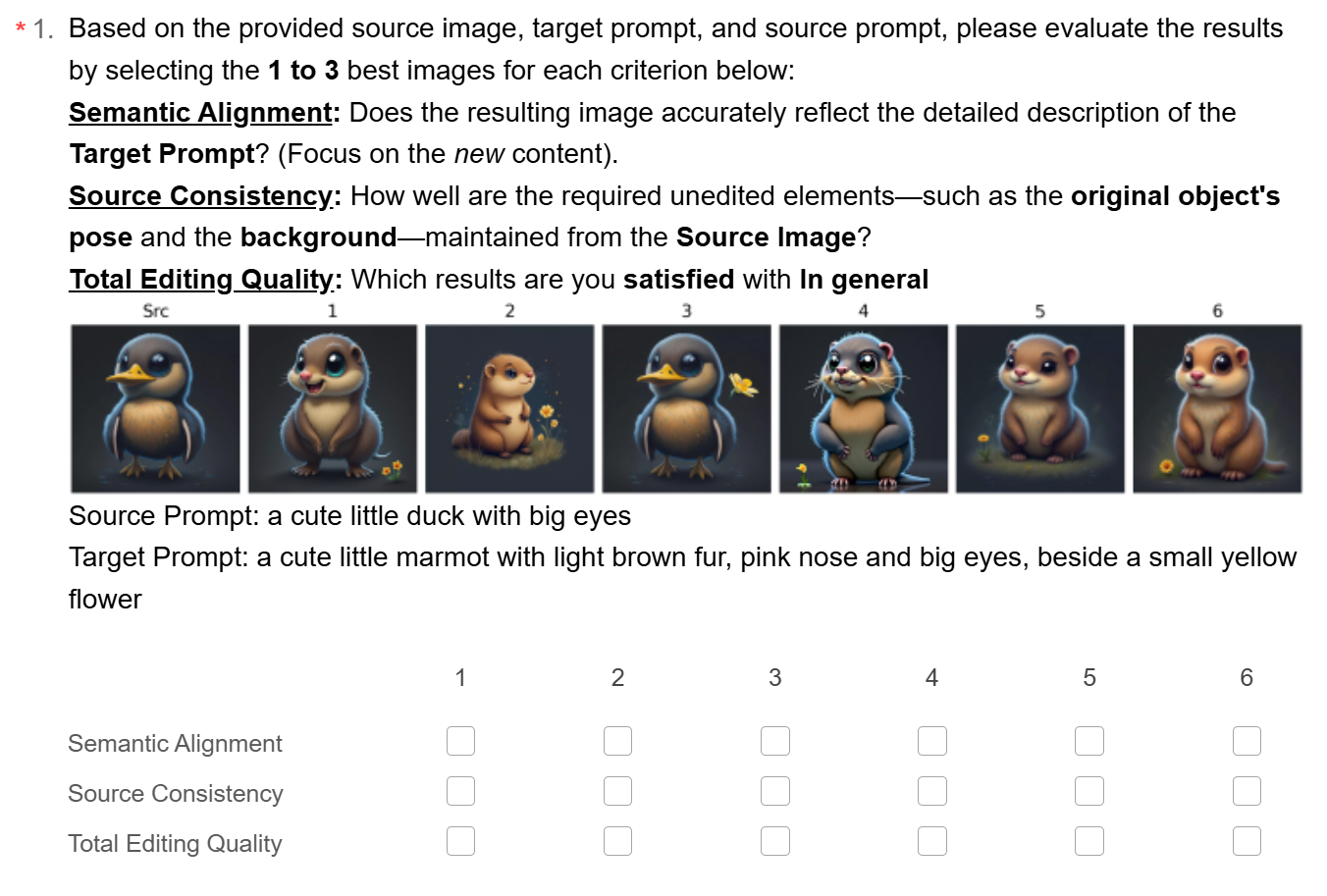}
    \caption{Example in user-study.}
    \label{fig:user_study}
    \vspace{-0.5em}
\end{figure}

\noindent\textbf{Setting.} 
We conducted a user study to compare our method against five baselines: FlowEdit~\cite{kulikov2025flowedit}, RF-Inversion~\cite{rf_inversion}, RF-Edit~\cite{rf-solver}, Multi-Turn~\cite{zhou2025multi}, and ParallelEdit~\cite{huang2024paralleledits}. 
The study comprised 15 trials, where each trial presented participants with a reference image, a source prompt, a target prompt and six target images. 
A total of 16 participants were invited to evaluate the results. In every trial, participants were asked to select one to three images based on three criteria: semantic alignment, source consistency, and total editing quality. 
For each method, the preference rate reflects the selection frequency aggregated across all trials and participants. 
The evaluation interface is illustrated in Fig.~\ref{fig:user_study}.

% For each trial, a target image was considered validated by human judgment if it received more than 2/3 of the total votes.

\noindent\textbf{Results.} 
 As reported in Table.~\ref{tab:user_study}, our method outperforms competing approaches across all metrics, achieving the highest scores in Semantic Alignment (SA), Source Consistency (SC), and Total Editing Quality (EQ).
% response_16
 \begin{table}[h]
    \centering
    \renewcommand{\arraystretch}{0.9}
    % \setlength{\aboverulesep}{2pt}
    % \setlength{\belowrulesep}{2pt}
    % \vspace{-0.5em}
    \caption{User study results of preference rates (\%) for Semantic Alignment (SA), Source Consistency (SC) and Total Editing Quality (EQ).}
    % \caption{User study results.}
    % \vspace{-0.5em}
    \label{tab:user_study} 
    \resizebox{0.7\columnwidth}{!}{%
    \begin{tabular}{lcccccc}
        % \hline
        \hline
        \textbf{Method} & Ours & FlowEdit & RF Edit & RF Inv & ParallelEdit & Multi Turn\\
        \midrule
         SA & \textbf{78.67} & 34.67 & 58.22 & 14.22 & 12.89 & 3.11 \\
         SC & \textbf{77.78} & 30.22 & 12.44 & 44.44 & 21.33 & 16.44 \\
         EQ & \textbf{76.44} & 19.11 & 20.44 & 18.22 & 7.56 & 3.11 \\
        \hline
        % \hline
    \end{tabular}
    }
    % \vspace{-0.5em}
\end{table}

\section{Editing Trajectory Analysis}
\label{sec:trajectory_analysis}
\begin{figure}[htbp]
    \centering
    \includegraphics[width=1.0\linewidth]{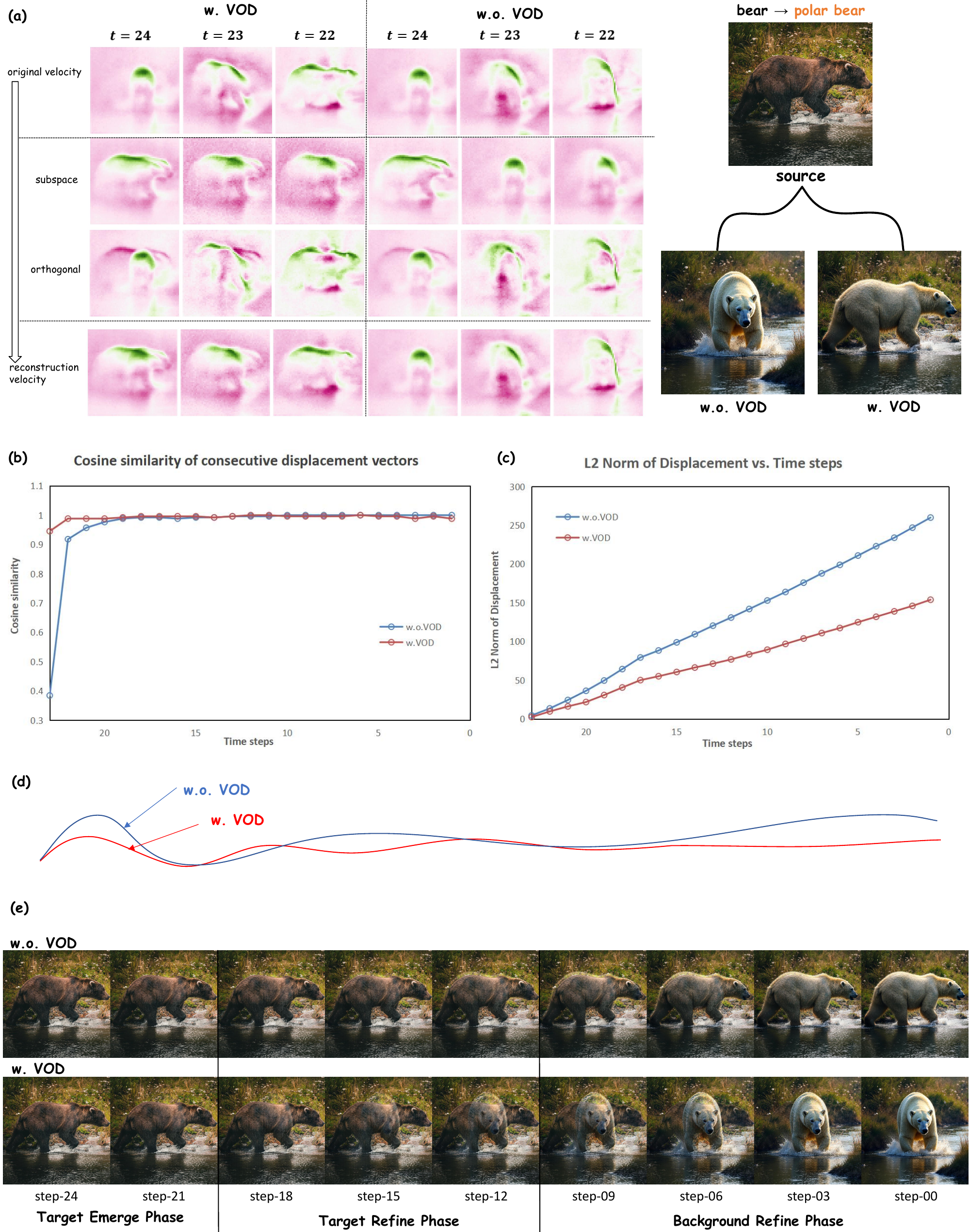}
    \caption{Visualization of the editing trajectory.}
    \label{fig:appendix_trajectory}
\end{figure}
As illustrated in Fig.~\ref{fig:ablation_example} and Fig.~\ref{fig:vod_ablation}, the orthogonal component typically corresponds to unstable structural changes that are irrelevant to the editing objectives.
By employing Velocity Orthogonal Decay (VOD), we effectively suppress this orthogonal component, thereby preserving source consistency. 
We further evaluate the application of VOD to single-target editing with the start time step set to $t_1 = 24/28$. As shown in Fig.~\ref{fig:appendix_trajectory}~(a), VOD successfully maintains source consistency in this setting as well.

In Fig.~\ref{fig:appendix_trajectory}~(b), we visualize the cosine similarity of consecutive displacement vectors. The similarity is notably low during the early stages without VOD, indicating an unstable editing trajectory. 
In contrast, with the application of VOD, the cosine similarity approaches 1.0, resulting in a significantly more stable trajectory. 
As demonstrated in Fig.~\ref{fig:appendix_trajectory}~(c), this stabilized trajectory incurs a lower transportation cost, which ultimately leads to superior source consistency. 
We provide a schematic abstraction of these two trajectories in Fig.~\ref{fig:appendix_trajectory}~(d).

Furthermore, by decoding the latent at each time step, as shown in Fig.~\ref{fig:appendix_trajectory}~(e), we can decompose the editing process into three distinct phases: \textit{target emergence}, \textit{target refinement}, and \textit{background refinement}. 
Initially, the high-level concept of the target (e.g., 'polar bear') manifests during the \textit{target emergence} phase. This is followed by the refinement of the subject. Finally, the model refines the overall image, including the background surrounding the target. 
Crucially, we observe that disruptions to source consistency (e.g., alterations in the bear's pose) originate during the target emergence phase and are subsequently refined into the final output.

\section{Hyperparameter Exploration for FlowDC}
\label{sec:flowdc_param}

\begin{figure}[h]
    \centering
    \includegraphics[width=0.6\linewidth]{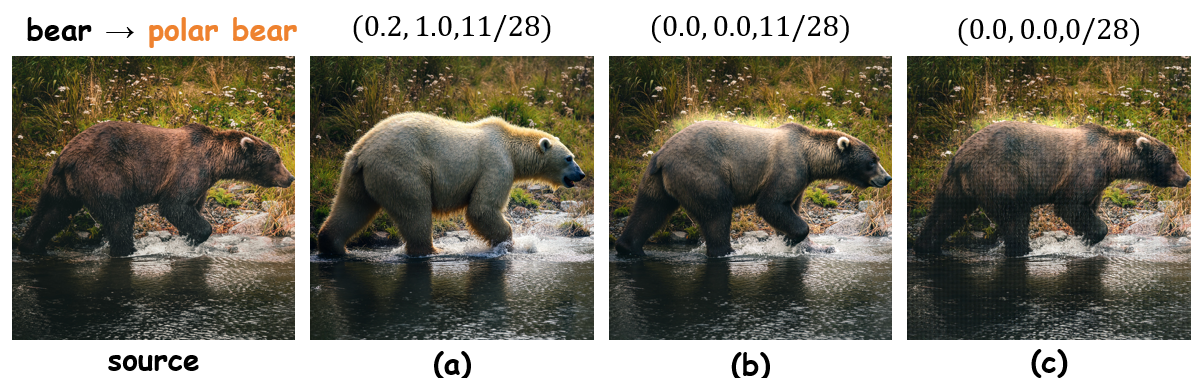}
    \caption{Ablation study on VOD decay hyperparameters. The tuple above each image denotes $(\lambda_1, \lambda_d, t_d)$.}
    \label{fig:vod_param}
    % \vspace{-0.5em}
\end{figure}

\noindent\textbf{Hyperparameters for VOD.} 
Given that significant structural deviations primarily arise during the first two phases, we concentrate the application of VOD on these stages. 
Moreover, recognizing that earlier time steps are more prone to compromising source consistency, we employ a decaying strategy where the decay strength starts at a high value and gradually diminishes over time.
We further analyze the impact of different VOD decay hyperparameters in Fig.~\ref{fig:vod_param}. 
As observed in Fig.~\ref{fig:vod_param}~(b), when the orthogonal component is completely suppressed (decayed to 0) only in the early phase, the model fails to effectively refine the 'polar bear' target. 
Conversely, extending this complete suppression throughout the entire process, as shown in Fig.~\ref{fig:vod_param}~(c), introduces visible grid artifacts in the final image.
These empirical observations motivate the specific parameter configurations detailed in Table~\ref{tab:hyperpara_ours}.

\noindent\textbf{Hyperparameters for PSO.} 
The parameter $t_o$ controls the operational steps of PSO. In practice, we configure PSO to run for a single step by setting $t_o = 27/28$, which is identical to $t_1$. 
Table~\ref{tab:quantitative_ablation} demonstrates that this single-step application is sufficient to improve semantic alignment. 
We further explore the impact of various PSO hyperparameters using the first 300 samples of Complex-PIE-Bench, as detailed in Table~\ref{tab:pso_param}. 
We observe that increasing the number of PSO steps from 1 to 7 leads to a deterioration in both semantic alignment and source consistency. 
This is because the initial PSO step is derived from velocities at the designated guidance time ($t_g$), which ensures high precision due to the low interpolation weight of Gaussian noise. 
In contrast, subsequent PSO steps rely on displacements at the current time step. These displacements contain minor biases compared to the precise editing direction, causing errors of orthogonal basis to accumulate over multiple orthogonal iterations of PSO. 
To ensure precision, we compute the guidance velocities by averaging the results of three calculations at the designated guidance time.
\begin{table}[h]
    \centering
    % \caption{Ablation on PSO steps evaluated on a subset (300 samples) of Complex-PIE-Bench.}    
    \caption{Ablation study on PSO steps.}    
    % \vspace{-0.5em}
    \label{tab:pso_param}
    \resizebox{0.55\columnwidth}{!}{%
    % \scriptsize
    \setlength{\tabcolsep}{3pt}
    \begin{tabular}{lcccc}
        \hline
        % \hline
        \textbf{PSO steps} & \textbf{CLIP-T (\%)} $\uparrow$ & \textbf{CLIP-I (\%)} $\uparrow$ & \textbf{DINO (\%)} $\uparrow$ & \textbf{LPIPS (\%)} $\downarrow$ \\
        \toprule
         \textbf{0} & 27.32 & \textbf{87.11} & \textbf{68.44} & \underline{24.62} \\
         \textbf{1} & \textbf{27.84} & \underline{86.76} & \underline{68.17} & \textbf{24.46} \\
         \textbf{7} & \underline{27.58} & 86.64 & 67.63 & 25.21 \\
        
        % \hline
        \hline
    \end{tabular}
    }
    % \vspace{-0.5em}
\end{table}

\section{Limitation}
\label{sec:limitation}
The main limitation of FlowDC lies in the underutilization of the PSO mechanism. Currently, PSO is restricted to a single step, as extending it to multiple steps introduces accumulated errors from biased displacement estimates. This constraint limits the method's potential in handling extremely complex editing scenarios that require sustained semantic guidance. Future work will focus on mitigating these displacement errors to enable stable multi-step PSO, thereby enhancing the controllability for intricate editing tasks.

\section{Full Algorithm}
\label{sec:full_algorithm}
Algorithm~\ref{pseudocode:full_algorithm} outlines the complete pipeline of FlowDC.

\begin{algorithm}[tb]
    \caption{Overall Pipeline of FlowDC}
    \label{pseudocode:full_algorithm}
\begin{algorithmic}
    \STATE {\bfseries Input:} 
    source image $X^{src}$, 
    source prompt $P^{src}$, 
    complex target prompt $P^{tar}$, 
    time steps $t_1, t_g, t_o$
    \STATE {\bfseries Output:} edited image $Z_0$ 
    \STATE $\{P^{tar_i}\}_{i=1}^n \gets LLM(P^{src}, P^{tar})$
    \STATE $\{Z^i_{t_1}\}_{i=1}^n \gets \{X^{src}\}_{i=1}^n$
    \STATE $\{v^i_{g}\}_{i=1}^n \gets \frac{1}{3} \sum_{k=1}^3 \mathrm{PVG}( \{P^{tar_i}\}_{i=1}^n, t_g)$
    % \STATE $\{v^i_{g}\}_{i=1}^n \gets PVG( \{P^{tar_i}\}_{i=1}^n, t_g)$
    \FOR{$t: t_1 \rightarrow t_o$}
        \FOR{$i: 1 \rightarrow n$}
        \STATE $v^i(t) \gets PVG(P^{tar_i}, t)$ \COMMENT{original}
        \STATE $d^i \gets Z^i_t - X^{src}$
        \IF{$t = t_1$} %\tikzmark{start-pso}
            \STATE $\mathbf{U}_i(t) \gets PVO(\{v^i_g\}_{j=1}^i)$
        \ELSE
            \STATE $\mathbf{U}_i(t) \gets PVO(\{d^j\}_{j=1}^i)$  
        \ENDIF %\tikzmark{end-pso}
        \STATE $v_{sub}^i(t) \gets Proj(v^i(t), \mathbf{U}_i(t))$
        \STATE $v_{orth}^i(t) \gets v^i(t) - v_{sub}^i(t)$
        \STATE $v^{i}{'}(t) \gets v_{sub}^i(t) + \lambda_{orth}(t)v_{orth}^i(t)$ \COMMENT{precise} 
        \STATE $Z^i_t \gets Z^i_t - v^{i}{'}(t)\Delta t$
        \ENDFOR
    \ENDFOR
    \STATE $Z_t \gets Z^n_t$
    \FOR{$t: t_1 \rightarrow t_o$}
        \STATE $v(t) \gets PVG(P^{tar}, t)$ \COMMENT{original}
        \STATE $d \gets Z_t - X^{src}$
        \STATE $\mathbf{U}(t) \gets d$ 
        \STATE $v_{sub}(t) \gets Proj(v(t), \mathbf{U}(t))$
        \STATE $v_{orth}(t) \gets v(t) - v_{sub}(t)$
        \STATE $v{'}(t) \gets v_{sub}(t) + \lambda_{orth}(t)v_{orth}(t)$ \COMMENT{precise} 
        \STATE $Z_t \gets Z_t - v{'}(t)\Delta t$
    \ENDFOR
    
\end{algorithmic}
\end{algorithm}

\section{Mathematical Justification of PSO}
\label{sec:mathematical_pso}
In this section, we mathematically verify why PSO enhances editability in FlowDC. Let $\mathcal{U}_{\text{PSO}} = \{u_i\}_{i=1}^n$ denote the subspace spanned by the orthogonal basis derived from the cumulative displacements $\{d_i\}_{i=1}^n$, and let $\mathcal{U}_{n} = \{d_n\}$ be the subspace spanned solely by the final displacement $d_n$. Defining $\mathcal{P}_{\text{PSO}}(v)$ and $\mathcal{P}_{n}(v)$ as the orthogonal projections of a velocity field $v$ onto $\mathcal{U}_{\text{PSO}}$ and $\mathcal{U}_{n}$ respectively, we assert:
\begin{equation}
    \label{equ:mathematical_pso}
    \|\mathcal{P}_{\text{PSO}}(v)\|^2 \ge \|\mathcal{P}_{n}(v)\|^2
\end{equation}
The equality holds if and only if all intermediate editing directions are collinear with the final direction $d_n$.

\noindent\textbf{Hilbert Projection Theorem}. Let $H$ be a \textbf{Hilbert space} and let $M$ be a closed \textbf{subspace} of $H$. For every vector $x \in H$, there exists a unique element $y \in M$ such that:
\begin{equation}
    \label{equ:hilbert_project}
    \| x - y \| \le \| x - z \|, \quad \langle x - y, z \rangle = 0, \quad \forall z \in M
\end{equation}

\noindent\textbf{Proof of Eq.~\ref{equ:mathematical_pso}}.
First, let $\mathcal{U}_{\text{latent}}$ denote the entire latent feature space. The subspaces $\mathcal{U}_{n}$ and $\mathcal{U}_{\text{PSO}}$ are defined as described previously. Since the final displacement $d_n$ is a component of the generating set $\{d_1, \dots, d_n\}$ of $\mathcal{U}_{\text{PSO}}$, it follows that $d_n \in \mathcal{U}_{\text{PSO}}$. Consequently, we establish the following nested subspace relationship:
\begin{equation}
\label{equ:subspace_inclusion}
    \mathcal{U}_{n} \subseteq \mathcal{U}_{\text{PSO}} \subseteq \mathcal{U}_{\text{latent}}
\end{equation}
Given a vector $v \in \mathcal{U}_{\text{latent}}$, according to the Hilbert Projection Theorem, $v$ can be decomposed into its projection onto $\mathcal{U}_{\text{PSO}}$ and an orthogonal residual $w_1$:
\begin{equation}
    \label{equ:project_v}
    \begin{split}
    &v = \mathcal{P}_{\text{PSO}}(v) + w_1, \\
    &\langle w_1, z \rangle = 0, \quad \forall z \in \mathcal{U}_{\text{PSO}}
    \end{split}
\end{equation}
Since $\mathcal{U}_{n} \subseteq \mathcal{U}_{\text{PSO}}$, $w_1$ is also orthogonal to $\mathcal{U}_{n}$:
\begin{equation}
    \label{equ:orthogonal_relation}
    \langle w_1, z \rangle = 0, \quad \forall z \in \mathcal{U}_{n}
\end{equation}
Next, we expand the projection of $v$ onto the smaller subspace $\mathcal{U}_{n}$. Using the linearity of the projection operator and the result from Eq.~\ref{equ:orthogonal_relation}:
\begin{equation}
    \label{equ:rewrite_project}
    \begin{aligned}
        \mathcal{P}_{n}(v) 
        &= \mathcal{P}_{n}(\mathcal{P}_{\text{PSO}}(v) + w_1) && (\text{by Eq.~\ref{equ:project_v}}) \\
        &= \mathcal{P}_{n}(\mathcal{P}_{\text{PSO}}(v)) + \mathcal{P}_{n}(w_1) && (\text{by Linearity}) \\
        &= \mathcal{P}_{n}(\mathcal{P}_{\text{PSO}}(v)) && (\text{since } w_1 \perp \mathcal{U}_n)
    \end{aligned}
\end{equation}
Now, we consider the projection of the vector $\mathcal{P}_{\text{PSO}}(v)$ onto $\mathcal{U}_{n}$. We can decompose $\mathcal{P}_{\text{PSO}}(v)$ into the component in $\mathcal{U}_{n}$ and a residual $w_2$ orthogonal to $\mathcal{U}_{n}$:
\begin{equation}
    \label{equ:decompose_pso_project}
    \begin{aligned}
        \mathcal{P}_{\text{PSO}}(v) 
        &= \mathcal{P}_{n}(\mathcal{P}_{\text{PSO}}(v)) + w_2 && (\text{by Eq.~\ref{equ:hilbert_project}})\\
        &= \mathcal{P}_{n}(v) + w_2 && (\text{by Eq.~\ref{equ:rewrite_project}})
    \end{aligned}
\end{equation}
where $\langle \mathcal{P}_{n}(v), w_2 \rangle = 0$ due to orthogonality. Finally, we obtain the norm relationship:
\begin{equation}
    \label{equ:final_proof}
    \|\mathcal{P}_{\text{PSO}}(v)\|^2 = \|\mathcal{P}_{n}(v)\|^2 + \|w_2\|^2
\end{equation}
Since the squared norm $\|w_2\|^2 \ge 0$, we conclude that $\|\mathcal{P}_{\text{PSO}}(v)\|^2 \ge \|\mathcal{P}_{n}(v)\|^2$.

\end{document}